\documentclass[fleqn,10pt]{wlscirep}
\usepackage[utf8]{inputenc}
\usepackage[T1]{fontenc}
\usepackage{epsfig}
\usepackage{epstopdf}
\usepackage{lineno}
\usepackage{amsmath}
\usepackage{setspace} 
\usepackage{tikz}
\usepackage{helvet}
\usepackage{caption}
\usepackage{amssymb}
\usepackage{bbding}
\usepackage{epstopdf}
\usepackage{graphicx}   
\usepackage{adjustbox}
\usepackage[absolute,overlay]{textpos}

\newcommand{\ra}[1]{\renewcommand{\arraystretch}{#1}}
\newcommand{\co}[1]{\setlength\tabcolsep{#1}}

\newcommand{\abbrfigure}{%
	\captionsetup[figure]{name=Fig.}
}

\newcommand{\beginextended}{%
	\setcounter{table}{0}
	\setcounter{figure}{0}
}

\newcommand{\extendeddatafigure}{%
	\captionsetup[figure]{name=Extended Data Fig.}
}
\newcommand{\extendeddatatable}{%
	\captionsetup[table]{name=Extended Data Table}
}

\newcommand{\beginsupplement}{%
	\setcounter{table}{0}
	\setcounter{figure}{0}
}
\newcommand{\supplementfigure}{%
	\captionsetup[figure]{name=Supplementary Fig.}
}

\newcommand{\supplementtable}{%
	\captionsetup[table]{name=Supplementary Table}
}

\title{Human-like object concept representations emerge naturally in multimodal large language models}

\author[1,2]{Changde Du}
\author[1,2]{Kaicheng Fu}
\author[3]{Bincheng Wen}
\author[1,2]{Yi Sun}
\author[1,2]{Jie Peng}
\author[1]{Wei Wei}
\author[1]{Ying Gao}
\author[1]{Shengpei Wang}
\author[1]{Chuncheng Zhang}
\author[4]{Jinpeng Li}
\author[1]{Shuang Qiu}
\author[3]{Le Chang}
\author[1,2,5,*]{Huiguang He}

\affil[1]{State Key Laboratory of Brain Cognition and Brain-inspired Intelligence Technology, Institute of Automation, Chinese Academy of Sciences, Beijing, China}
\affil[2]{School of Artificial Intelligence, University of Chinese Academy of Sciences, Beijing, China} \affil[3]{Institute of Neuroscience, State Key Laboratory of Brain Cognition and Brain-Inspired Intelligence Technology, CAS Center for Excellence in Brain Science and Intelligence Technology, Chinese 
Academy of Sciences, Shanghai, China}
\affil[4]{School of Automation Science and Engineering, South China University of Technology, Guangzhou, China}
\affil[5]{Zhongguancun Academy, Beijing, China} 

\affil[*]{corresponding author: Huiguang He (huiguang.he@ia.ac.cn)}

\begin{abstract}
	Understanding how humans conceptualize and categorize natural objects offers critical insights into perception and cognition. With the advent of Large Language Models (LLMs), a key question arises: can these models develop human-like object representations from linguistic and multimodal data? In this study, we combined behavioral and neuroimaging analyses to explore the relationship between object concept representations in LLMs and human cognition. We collected 4.7 million triplet judgments from LLMs and Multimodal LLMs (MLLMs) to derive low-dimensional embeddings that capture the similarity structure of 1,854 natural objects. The resulting 66-dimensional embeddings were stable, predictive, and exhibited semantic clustering similar to human mental representations. Remarkably, the dimensions underlying these embeddings were interpretable, suggesting that LLMs and MLLMs develop human-like conceptual representations of objects. Further analysis showed strong alignment between model embeddings and neural activity patterns in brain regions such as EBA, PPA, RSC, and FFA. This provides compelling evidence that the object representations in LLMs, while not identical to human ones, share fundamental similarities that reflect key aspects of human conceptual knowledge. Our findings advance the understanding of machine intelligence and inform the development of more human-like artificial cognitive systems.
\end{abstract}
\begin{document}
	
	\flushbottom
	\maketitle
	\abbrfigure
	
	\section*{Introduction}
	
	The ability to categorize and conceptualize objects forms the bedrock of human cognition, influencing everything from perception to decision-making. When confronted with diverse objects, humans can often differentiate their categories and concepts by making structured comparisons between them. This process is an essential part of human cognition in tasks ranging from everyday communication to problem-solving. In this cognitive process, our mental representations serve as a substrate, aiding in the recognition of objects \cite{biederman1987recognition,edelman1998representation}, formation of categories \cite{nosofsky1986attention,goldstone1994role,rosch1976basic}, organization of conceptual knowledge \cite{mahon2009concepts,rogers2004semantic}, and the prediction of behaviors based on experiences. Therefore, understanding the structure of these representations is a fundamental pursuit in cognitive neuroscience and psychology \cite{shepard1987toward,battleday2020capturing,jagadeesh2022texture,grand2022semantic}, underpinning significant research advancements in the field. For instance, various studies have identified potential dimensions that organize these representations, such as animals versus non-animals \cite{connolly2012representation,downing2006domain,kriegeskorte2008matching,caramazza1998domain}, natural versus human-made \cite{hebart2020revealing,hebart2023things}, and large versus small \cite{konkle2012real,konkle2011canonical}.

	The cognitive plausibility of deep learning systems has sparked significant debate \cite{bowers2023deep,hermann2023for}, with recent works often focusing on diverse neural networks pretrained on limited datasets for specific computer vision tasks like image classification \cite{jha2023extracting,nadler2023divergences,cohen2020separability,dobs2022brain,mahner2024dimensions,jacob2021qualitative}. 
	While these endeavors have led to notable advancements \cite{jacob2021qualitative,goldstein2022shared,muttenthaler2022interpretable,saxe2021if}, including some evidence of human-like representations emerging from self-supervised learning \cite{prince2024contrastive,konkle2022self,zhuang2021unsupervised,feather2023model}, a critical question remains: to what extent can complex, task-general psychological representations emerge without explicit task-specific training, and how do these compare to human cognitive processes across a broad range of tasks and domains?
	LLMs, such as OpenAI's ChatGPT and Google's Gemini, have emerged as potent tools in text and image understanding, generation, and reasoning. These models exhibit impressive capabilities in tasks like object identification, information categorization, concept communication, and inference.  Unlike task-specific small-scale neural network models, LLMs utilize generic neural network architectures with billions of parameters, trained through next token prediction on massive text corpora (and images for MLLMs) comprising trillions of tokens. Despite ongoing debates about their capacities \cite{demszky2023using,dillion2023can,messeri2024artificial}, one potential strength lies in their adeptness at problem-solving with minimal task-specific training, often requiring only straightforward task instructions without parameter updates. These features raised the question of whether LLMs have developed human-like conceptual representations about natural objects.

	In this study, we used a data-driven approach to explore the core dimensions of mental representations in LLM (ChatGPT-3.5) and MLLM (Gemini Pro Vision 1.0). Inspired by previous work conducted on human similarity judgments using visual object images, we adopted a similar methodology to both the LLM and MLLM. Unlike presenting visual stimuli to human participants and MLLMs, we presented corresponding textual descriptions of visual images to the LLMs. Harnessing the models' ability to perform a triplet odd-one-out task, a well-established paradigm in cognitive psychology \cite{hebart2020revealing,josephs2023dimensions,hebart2023things,jagadeesh2022texture}, we collected extensive datasets comprising 4.7 million triplet similarity judgments for both the LLM and MLLM. Each dataset is rich in triple similarity judgment entries, drawn from a pool of 1,854 unique objects. This diverse collection enables the examination and capture of visual and conceptual mental representations spanning a wide array of natural objects.
	
	Using a representation learning method previously designed for human participants \cite{zheng2019revealing,hebart2020revealing}, we identified 66 sparse, non-negative dimensions underlying LLMs' similarity judgments that lead to excellent predictions of both single-trial behavior and similarity scores between pairs of objects. We demonstrated that these dimensions are interpretable, exhibited spontaneous semantic clustering, and characterized the large-scale structure of LLMs' mental representations of natural objects. Furthermore, by comparing the identified dimensions with the core dimensions observed in human cognition, we found close alignment between model and human embeddings. Finally, we found strong correspondence between the model embeddings and neural activity patterns in category-selective brain Region of Interests (ROIs, e.g., EBA, PPA, RSC, FFA), underscoring the generalization of these learned mental representations and offering a compelling evidence that the object representations in LLMs, while not identical to those in the human, share fundamental commonalities that reflect key schemas of human conceptual knowledge. These results enrich the growing body of work characterizing the emergent characteristics of LLMs \cite{binz2023using,webb2023emergent,wei2022emergent,schaeffer2024emergent,hagendorff2023machine,hagendorff2023human,strachan2024testing,kumar2024shared,chen2023emergence,zhang2024mathverse}, showcasing their potential to capture and reflect human-like conceptualizations of real-world objects.
	\begin{figure}[!htbp]
		\centering
		\includegraphics[width=17cm]{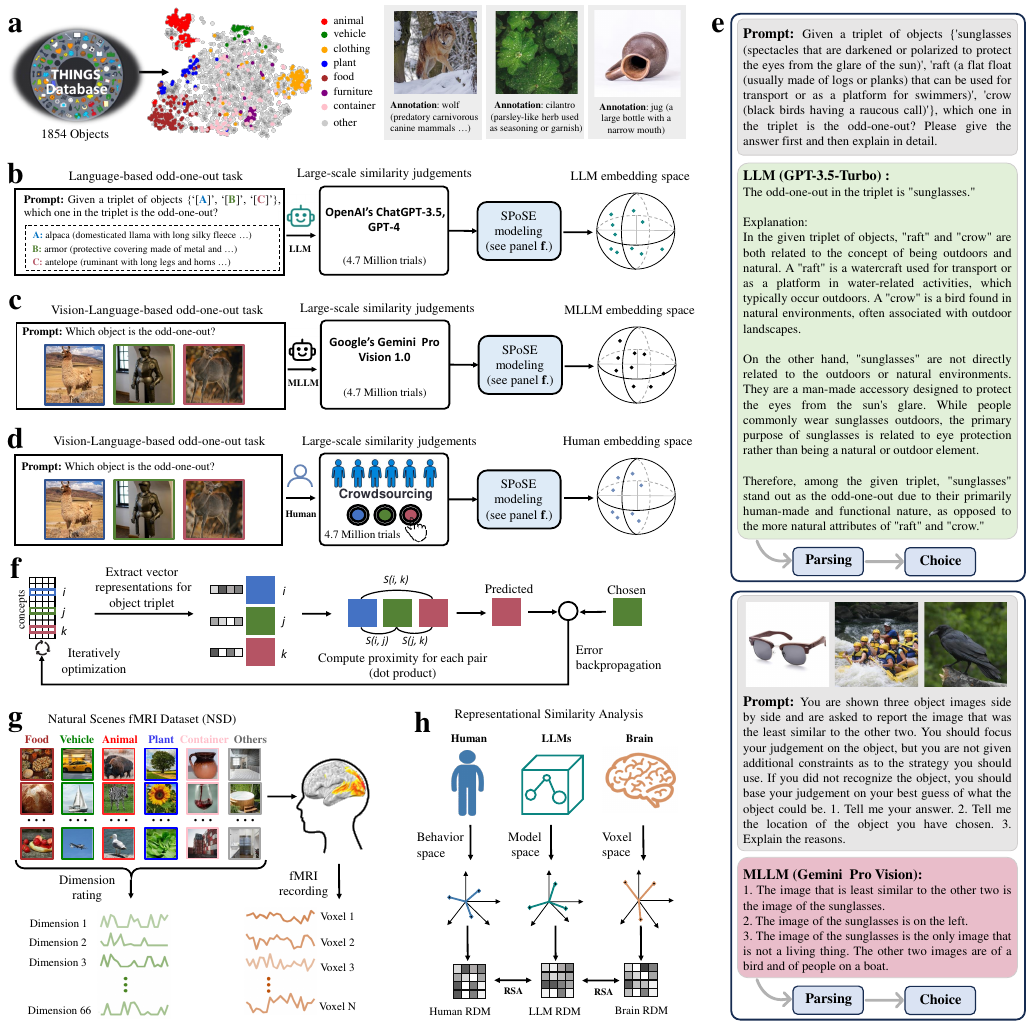}
		\caption{\textbf{Schematic diagrams of the experiment and analysis methods.} \textbf{a}, THINGS database and examples of object image with their language descriptions at the bottom. \textbf{b-d}, Pipelines of mental embedding learning under the triplet odd-one-out paradigm for LLM, MLLM, and humans, respectively. Odd-one-out judgments were collected for approximately 4.7 million triplets, and modeled using the SPoSE approach to derive the corresponding low-dimensional embedding. \textbf{e}, Examples of prompts and responses for LLM and MLLM. \textbf{f}, Illustration of the SPoSE modeling approach. \textbf{g}, Illustration of the NSD dataset with dimension ratings for stimulus images. The schematic structure incorporates elements adapted from Figure 1A of Horikawa et al. (2020)\cite{horikawa2020neural} (https://doi.org/10.1016/j.isci.2020.101060), published under a CC BY 4.0 license. \textbf{h}, Overview of the comparisons between space of LLMs, human behavior and brain activity. For this figure, all images were replaced by images with similar appearance from the public domain. Images used under a CC0 license, from Pixabay and Pexels.}
		\label{fig:main}
	\end{figure}
	
	\section*{Results}	
	We initiated our study by selecting a diverse set of objects from the THINGS database \cite{hebart2019things}, encompassing 1,854 common objects (Fig. \ref{fig:main}a). To compare LLMs' mental representations with humans, we adopted the triplet odd-one-out task, effective for modeling human mental dimensions \cite{hebart2020revealing,josephs2023dimensions,hebart2023things,jagadeesh2022texture,wei2024cocog} (Figs. \ref{fig:main}b-d).
	Given the impracticality of conducting 1.06 billion triplet judgments, we approximated the similarity matrix using approximately $0.44\%$ of the total judgments, following established methods \cite{hebart2020revealing,hebart2023things}. Human similarity judgments were collected from 4.7 million trials via Amazon Mechanical Turk \cite{hebart2023things}, and LLMs' behavioral data mirrored these trials. Fig. \ref{fig:main}e displays examples of prompts and responses from GPT-3.5-Turbo and Gemini Pro Vision, detailing choice derivation.
	We utilized the Sparse Positive Similarity Embedding (SPoSE) method \cite{zheng2019revealing,hebart2020revealing} (Fig. \ref{fig:main}f) to infer LLMs' low-dimensional representations, optimizing object weights to predict behavioral judgments. We validated the generalization of LLM embeddings on the Natural Scenes Dataset (NSD) \cite{allen2022massive} and applied Representational Similarity Analysis (RSA) \cite{kriegeskorte2008representational} to assess correlations with neural activity (Figs. \ref{fig:main}g-h).\\

	\noindent \textbf{Low-dimensional embeddings identified from LLMs are stable and predictive}	
	
\begin{figure}[!htbp]
		\centering
		\includegraphics[width=17cm]{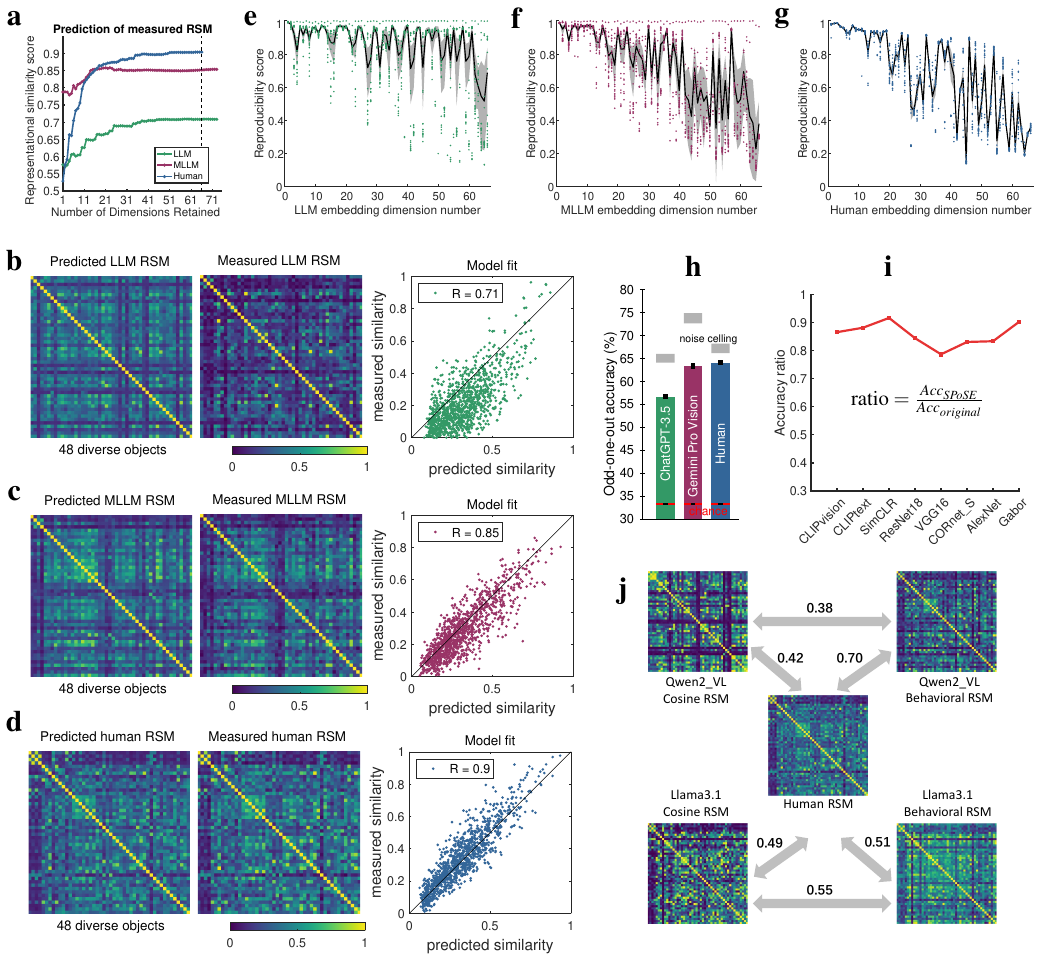}
		\caption{\textbf{Validation of the embeddings derived from similarity judgments over 4.7 million trials.} \textbf{a}, Prediction performance of the measured similarity matrix with varying dimensions of the SPoSE embedding. \textbf{b-d}, RSMs for a subset consisting of 48 objects, created by estimating similarity based on the model embedding (left) and by fully sampling all possible triplets in a validation behavioral experiment (middle). Here, the similarity between two objects is operationalized as the proportion of times they are judged to be similar, across all trials. Correlation between the predicted and measured similarity on all object pairs were shown in right. \textbf{e-g}, Reproducibility of dimensions in the chosen 66-dimensional embedding. The dimensions were sorted in descending order by the sum of their weights across objects. The scores are presented as mean $\pm$ 95\% confidence intervals (CIs), and shaded areas reflect the 95\% CIs (n=20 runs, and each dot represents the highest correlation of each selected dimension with all dimensions of a single run).  \textbf{h}, Odd-one-out prediction performance on the model's own held-out behavioral choice test set. Results and chance-levels are presented as mean $\pm$ 95\% CIs, and the error bars reflect 95\% CIs (n=1000 bootstraps). The noise ceilings were estimated from the additional behavioral datasets for each model separately, and were presented as mean $\pm$ 95\% CIs (shaded bands). \textbf{i}, How closely SPoSE embeddings mimic model's original features in odd-one-out predictions. The vertical axis represents the ratio of the SPoSE embedding accuracy to the original feature accuracy on the held-out test set constructed using cosine distances. \textbf{j}, How correlated are the model probing methods based on behavioral choices with those based on cosine distance. The numbers on the gray arrows represent the Pearson correlation between different RSMs (of the 48 objects).}
		\label{fig:66dimensional}
	\end{figure}

	Given the stochastic nature of SPoSE modeling (see Methods), we conducted multiple reruns with different random initializations, yielding slightly varied embeddings. Dimensions were sorted by their total object weights, and redundant dimensions (correlation > 0.4) were pruned, retaining only one. This reduced redundancy, as most dimensions appeared consistently across runs.  To evaluate retained dimensions, we gathered triplet judgments for 48 typical objects (these triplet judgments are not included in the SPoSE model's training data), comparing choice probabilities with predictions from the SPoSE embedding. Fig. \ref{fig:66dimensional}a shows that predictive performance stabilizes as dimensions increase, saturating at 60 dimensions for LLM, MLLM, and human. We chose the top 66 dimensions for LLM and MLLM to align with the 66 core dimensions from human similarity judgments \cite{hebart2023things}, as dimensions beyond the 66th contribute minimally to object similarity prediction.
	
	Figs. \ref{fig:66dimensional}b-d illustrate strong correlations between the model-predicted and behaviorally-measured Representational Similarity Matrices (RSMs) for LLM (0.71), MLLM (0.85), and human (0.9), validating the close reflection of behavioral similarity space. This result shows that, despite the complex object pool, a low-dimensional embedding can capture a large portion of the representational structure derived from similarity judgments.
	
	Next, we calculated reproducibility scores for each retained dimension (see Methods). In Fig. \ref{fig:66dimensional}e, all LLM embedding dimensions scored above 0.51, with 37 dimensions exceeding 0.90. Fig. \ref{fig:66dimensional}f shows that MLLM dimensions had reproducibility scores above 0.36, except one at 0.22, with 31 dimensions exceeding 0.80. Human dimensions in Fig. \ref{fig:66dimensional}g showed comparable reproducibility. These findings confirm that the embeddings are stable across reruns.
	
	We also evaluated the ability of these embeddings to predict choices in the odd-one-out task using model's own held-out behavioral choice test set. As shown in Fig. \ref{fig:66dimensional}h, accuracies were 56.7\% ($\pm$0.22\%), 63.4\% ($\pm$0.25\%), and 64.1\% ($\pm$0.18\%) for LLM, MLLM, and human, respectively (chance = 33.3\%, 95\% CI = [33.19\%, 33.47\%], 1,000 permutation tests). Noise ceilings for fitting individual-trial behavior were 65.1\% ($\pm$0.96\%), 73.8\% ($\pm$1.12\%), and 67.2\% ($\pm$1.04\%), indicating that the low-dimensional embeddings achieve up to 87.1\%, 85.9\%, and 95.4\% of the optimal predictive accuracy for LLM, MLLM, and human, respectively.
	
	Furthermore, we compared SPoSE embedding's predictive performance to that of the original model features using open-source models. As shown in Fig. \ref{fig:66dimensional}i, the accuracy ratios demonstrate that SPoSE embeddings closely approximate the original features (with ratios around 90\%), highlighting their effectiveness as compressed representations (see Extended Data Fig. 1a for the number of retained dimensions for these models and their predictive performance curves). Additionally, in Fig. \ref{fig:66dimensional}j, we compared two model probing methods: the behavioral judgment method and the cosine distance method. For the pure language model Llama3.1, the correlation between the two methods was relatively strong ($r = 0.55$), while for the vision-language model Qwen2\_VL \cite{wang2024qwen2}(7B version), it was lower ($r = 0.38$). Importantly, the behavioral judgment method aligned better with human-derived RSM than the cosine distance method (0.70 vs. 0.42 for Qwen2\_VL, and 0.51 vs. 0.49 for Llama3.1). These results suggest the feasibility of using SPoSE embeddings derived from behavioral judgments to probe the closed-source LLMs/MLLMs where direct feature extraction is infeasible.
	
	Overall, SPoSE modeling generated a low-dimensional, stable, and predictive mental embedding, excelling in predicting triplet similarity judgments and reconstructing their representational space. This indicates that LLM (particularly MLLM) judgments of natural objects are structured and principled. In the following sections, we explore key schemas in this embedding and their connections to human mental representations.\\

	\noindent \textbf{Emergent object category information}
	
	\begin{figure}[!htbp]
		\centering
		\includegraphics[width=17cm]{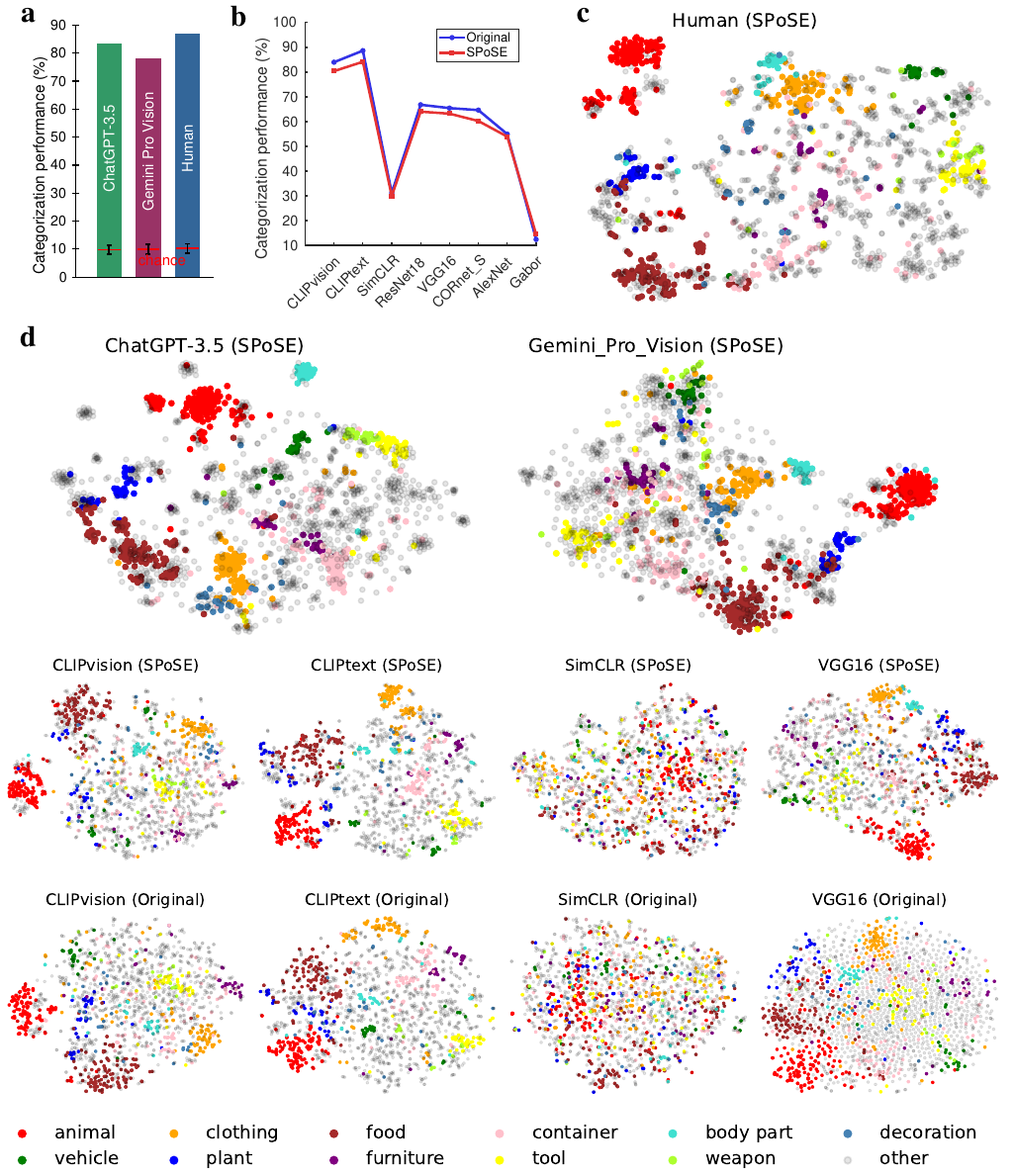}
		\caption{\textbf{Emerging object category information in the derived embeddings}. \textbf{a}, Categorization performance of different embeddings,  tested on 18 categories in the THINGS database. Chance-levels are presented as mean $\pm$ 95\% CIs, and the error bars reflect 95\% CIs (n=1000 bootstraps). \textbf{b}, Categorization performance comparisons between the SPoSE embedding and original model feature. \textbf{c-d}, t-SNE visualization of 1,854 objects, showing emergent category clusters in the learned embedding space of human and models. Dots correspond to objects, and were colored according to their labels.}
		\label{fig:tsne}
	\end{figure}

	Natural object categories emerge from mental embeddings derived from human similarity judgments \cite{hebart2020revealing,josephs2023dimensions}. To assess whether embeddings from LLM and MLLM also show emergent category structures, we used 18 high-level categories from the THINGS database \cite{hebart2019things} and applied a cross-validated nearest-centroid classifier to predict the category membership for each of the 1,112 objects of these categories (see Methods). 
	
	As seen in Fig. \ref{fig:tsne}a, LLM embeddings achieved 83.4\% top-1 accuracy (chance = 9.8\%, 95\% CI = [8.2\%, 11.4\%]), while MLLM reached 78.3\% (chance = 9.9\%, 95\% CI = [8.2\%, 11.5\%]). Human embeddings performed best with 87.1\% top-1 accuracy  (chance = 10.3\%, 95\% CI = [8.6\%, 12.0\%]). Fig. \ref{fig:tsne}b shows similar categorization performance between SPoSE embeddings and original features across models, confirming SPoSE's effectiveness in capturing object categories if the model itself is powerful in object representation \cite{cohen2020separability}.
	Figs. \ref{fig:tsne}c-d visualizes the global structure of embeddings via a t-SNE plot (dual perplexity: 5 and 30; 1,000 iterations) initialized with multidimensional scaling (MDS). Objects with similar values cluster together, showing that items from the same category group across LLM, MLLM, and human data. Thus, LLMs inherently capture object category structures without explicit representational constraints. Compared to traditional supervised models (like VGG16 \cite{Karen2015vgg}) or self-supervised models (like SimCLR \cite{chen2020simple}), LLMs and humans exhibit superior object category information.
	Overall, LLM and MLLM results support known distinctions between animate/inanimate and man-made/natural objects, consistent with previous human studies \cite{hebart2020revealing}. \\

	\noindent \textbf{The embedding dimensions of the LLMs are interpretable and informative}
	
\begin{figure}[!htbp]
		\centering
		\includegraphics[width=17cm]{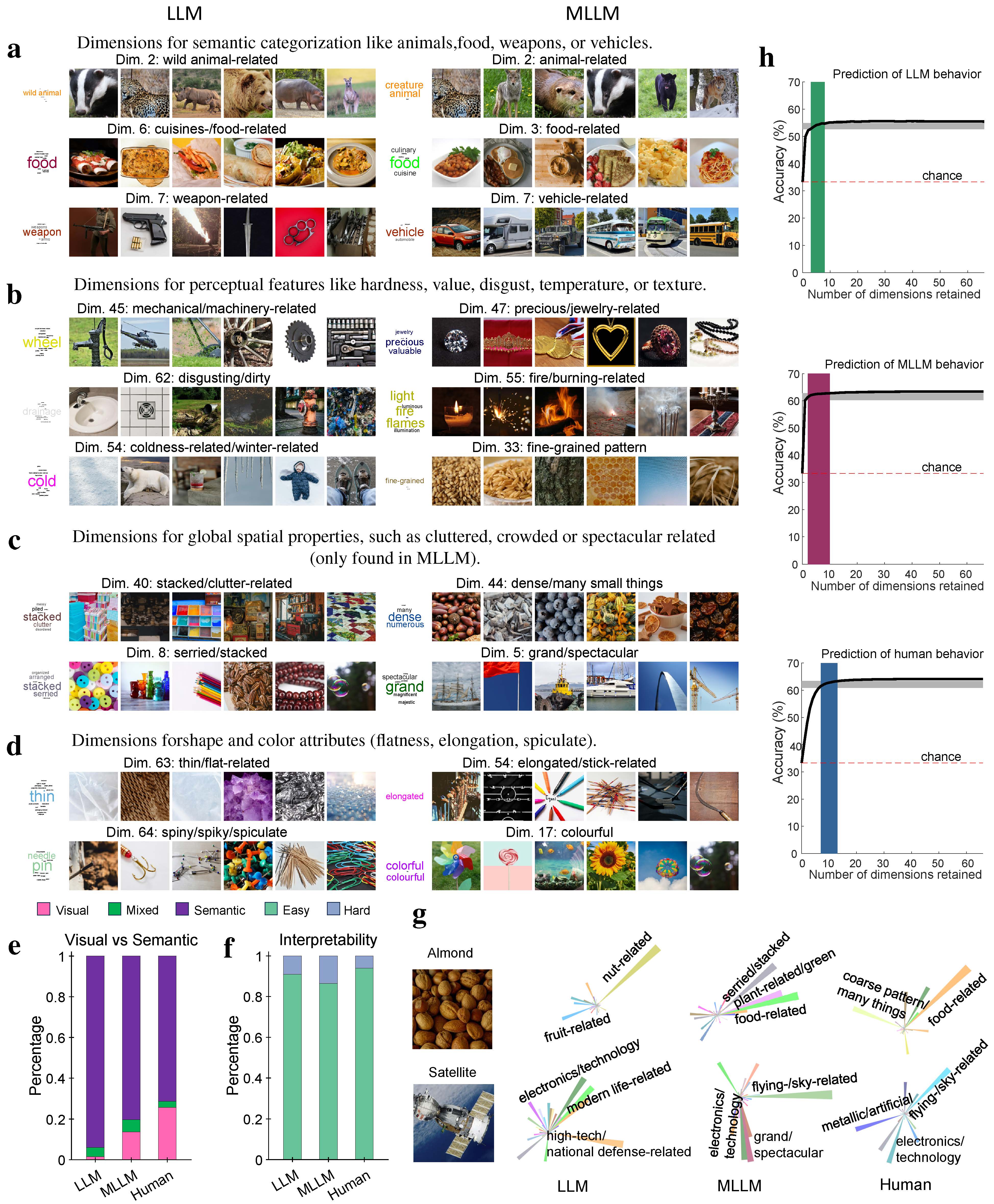}
		\caption{\textbf{Object dimensions illustrating their interpretability.} \textbf{a-d}, For each dimension, visualization includes the top 6 images carrying the greatest weights, accompanied by a word cloud reflecting human's annotations for what is captured by the dimension. For LLM, we replaced linguistic descriptions with images of the related objects to aid visualization. \textbf{e}, Proportions of visual, semantic, and mixed visual-semantic dimensions.  \textbf{f}, Proportions of easy and hard to interpret. \textbf{g}, Illustration of example objects with their dominant dimensions. \textbf{h}, To explain 95 to 99\% of the predictive performance in behavior, how many dimensions are required. For subfigures \textbf{a-d, g}, all images were replaced by images with similar appearance from the public domain. Images used under a CC0 license, from Pixabay and Pexels.}
		\label{fig:dim_vis}
	\end{figure}

	While past research has explored multidimensional mental representations in humans \cite{hebart2020revealing,hebart2023things}, this study is the first to examine LLMs. We focused on analyzing these dimensions to identify properties prioritized by LLM and MLLM when assessing object similarity.
	Figs. \ref{fig:dim_vis}a-d visually represent selected dimensions in LLM and MLLM by showing object images weighted most heavily in those dimensions. These dimensions are interpretable, reflecting conceptual and perceptual traits. We assigned intuitive labels (e.g., "animal-related" and "food-related"; see Methods) to dimensions from LLM and MLLM. Some dimensions appear to represent semantic categories (e.g., food, animals, vehicles) (Fig. \ref{fig:dim_vis}a), while others capture perceptual features like hardness, value, temperature, or texture (Fig. \ref{fig:dim_vis}b). Certain MLLM dimensions seem to reflect global spatial properties (e.g., crowded) (Fig. \ref{fig:dim_vis}c), while some convey shape (flatness, elongation) and color (Fig. \ref{fig:dim_vis}d). Dimensions also distinguish user specificity (children vs. adults, everyday consumers vs. experts) (Extended Data Fig. 1b), physical composition (wood, ceramic, metal) (Extended Data Fig. 1c), and environment-related traits (land vs. sea, indoor vs. outdoor) (Extended Data Fig. 1d). See Extended Data Figs. 2-6 for a visual display of all 66 dimensions. Each dimension in LLM or MLLM embodies multiple attributes, but we offer a single interpretation per dimension to showcase the concepts they represent. 
	
	We categorized the dimensions into three groups: shared across all three (LLM, MLLM, human), unique to human, and missing from human but present in LLM/MLLM. Shared dimensions include "animal-related" (2, 3), "food-related" (2, 3, 6, 18, 41, 58), "electronics/technology" (5, 11), "transportation/movement" (8, 19, 52, 58),	and more. 
	Unique human dimensions include 	"white" (22), "red" (24), "black" (27), "tubular" (31), "grid/grating-related" (33), "spherical/voluminous" (36), "elliptical/curved" (41),  and more.  
	Dimensions missing in humans but present in LLM/MLLM include "vegetable-related" (13, 28),
	"frozen treats/drink" (22),
	"presentation/display-related" (23),
	"headwear-related" (25),
	"livestock-related" (26),
	and more. 
	In general, categories such as animals, food, and technology are universally recognized across humans, LLMs, and MLLMs, indicating a common conceptual basis. Humans excel at distinguishing object differences through perceptual features like color, shape, and texture, which are less pronounced in LLM and MLLM. Moreover, LLM and MLLM tend to form more specific categories (e.g., fruits, vegetables, headwear) than humans' broader categorizations. The absence of certain dimensions in human representations does not imply an inability to perceive them; rather, these dimensions may emerge at a higher level, such as humans consolidating "vegetable-related" and "nut-related" dimensions under a "food-related" dimension.
	
	The dimensions derived from LLM and MLLM appear to exhibit a degree of interpretability, as evidenced by the ability to assign intuitive labels to them. These labels were listed in Extended Data Table 1. We also annotated these dimensions using MLLM, comparing human-generated vs. MLLM-generated labels in Extended Data Table 2.
	In addition, we divided all dimensions into visual, semantic, and mixed visual-semantic groups (based on examination by human experts) and calculated the proportion for each group (Fig. \ref{fig:dim_vis}e). LLM and MLLM have more semantic dimensions, while humans are better at using visual information. In contrast, the purely vision model SimCLR (a self-supervised learning model) shows minimal ability to learn semantic dimensions (Extended Data Fig. 7), whereas the dimensions derived from random representations lack any interpretability (Supplementary Fig. 1). We also categorized dimensions by ease of interpretation (based on whether they can be clearly explained by a single label), finding that most dimensions are easy to interpret (Fig. \ref{fig:dim_vis}f). Specifically, 60/66 dimensions for LLM, 57/66 for MLLM, and 62/66 for humans are easy to interpret, with humans having the fewest hard-to-interpret dimensions.
	
	We examined the composition of dimensions for specific objects. Fig. \ref{fig:dim_vis}g uses circular bar plots to represent objects, where petal angle and color denote dimensions, and length indicates the dimension's importance. For example, "almond" is primarily food-related, while "satellite" is associated with electronics and flying.  These plots also demonstrate that objects are indeed characterized by a rather small number of dimensions, indicating that not all 66 dimensions are necessary for particular similarity judgment.
	To quantify this, we progressively eliminated less significant dimensions for each object and assessed model performance. We found that retaining 3 to 8 dimensions for LLM, 2 to 10 for MLLM, and 7 to 13 for humans suffices to achieve 95-99\% of the full model's performance in explaining behavioral judgments within the odd-one-out context (Fig. \ref{fig:dim_vis}h). LLM exhibits lower dimensionality than humans, likely due to its lack of visual input. Although MLLM can access visual data, its multimodal integration remains inferior to human capabilities, limiting dimensions related to shape or color, inherently tied to human visual experience.\\

	\noindent \textbf{Comparison between models and humans}
	
	We employed two approaches to assess model-human alignment: one measuring consistency in similarity judgments \cite{rajalingham2018large} and the other analyzing core dimension relationships.
	
\begin{figure}[!htbp]
	\centering
	\includegraphics[width=17cm]{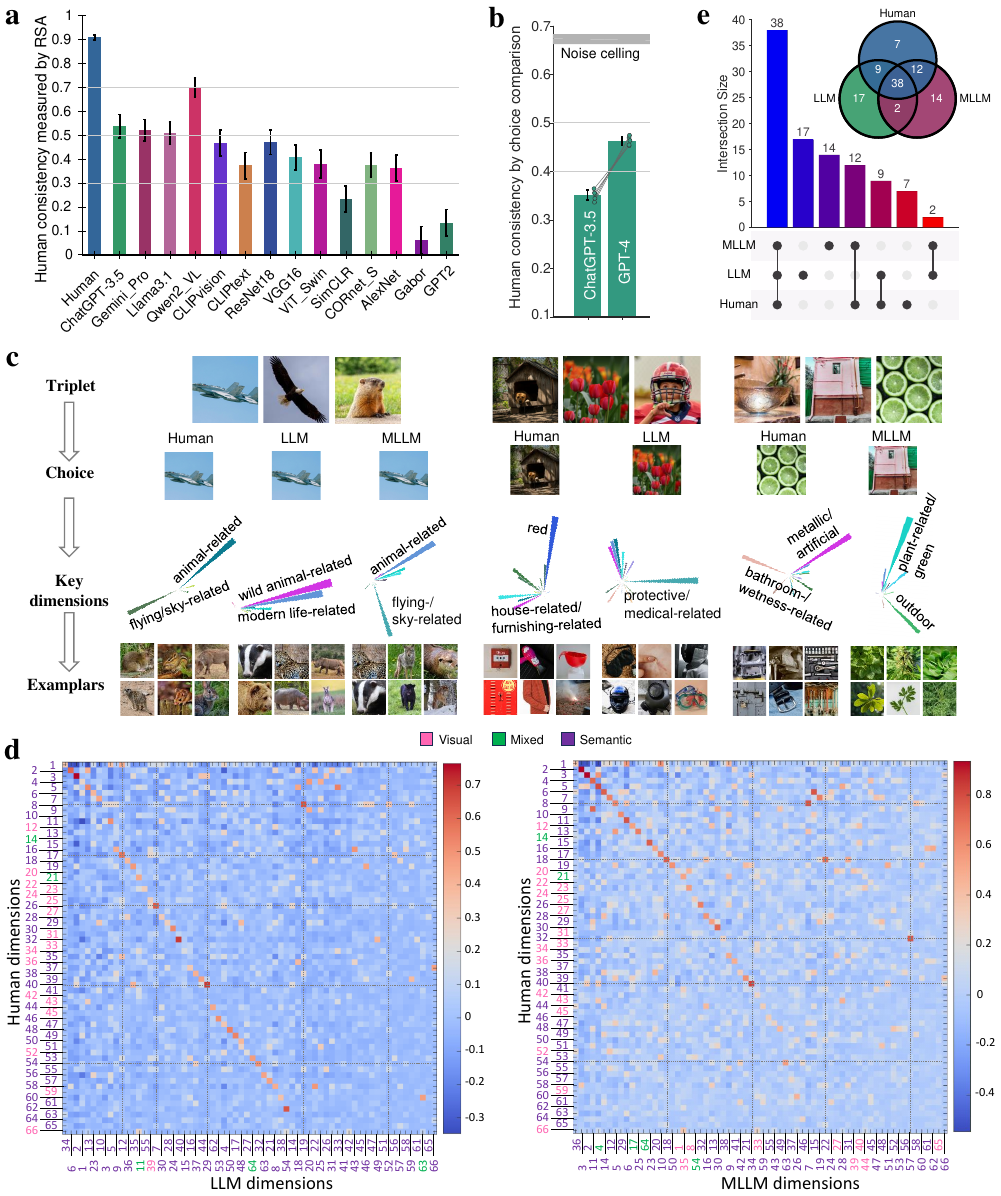}
	\caption{\textbf{Comparison between models and humans.} \textbf{a}, Human-model consistency (Pearson's $r$) between human and model object similarity matrices. Left blue bar shows baseline between-human consistency. Data are presented as mean $\pm$ 95\% CIs, and the error bars reflect 95\% CIs (n=1000 bootstraps).  \textbf{b}, Preliminary comparison between ChatGPT-3.5 and GPT-4. The error bars reflect standard deviation (SD), and data are presented as mean $\pm$ SD (n=5 samplings, and dots represent the result of each time).  \textbf{c}, Key dimensions that underpin specific behavioral choices made by human and models. \textbf{d}, Cross-correlation matrix between each pair of model systems (human-LLM, human-MLLM, and LLM-MLLM (in Extended Data Fig. 1e)). \textbf{e}, Quantification of shared ($r>0.2$) and non-shared dimensions between different systems. For subfigure \textbf{c}, all images were replaced by images with similar appearance from the public domain. Images used under a CC0 license, from Pixabay and Pexels.}
	\label{fig:correlation_matirx}
    \end{figure}

	Using comprehensive triplet sampling on 48 objects, we estimated similarity via choice probabilities and correlated model and human similarity matrices with Pearson correlation. Fig. \ref{fig:correlation_matirx}a compares various models, including visual-only, visual-language, LLMs, MLLMs, and a Gabor baseline, revealing higher human-consistency for LLM and MLLM.
	A preliminary comparison between ChatGPT-3.5 and GPT-4 in Fig. \ref{fig:correlation_matirx}b, directly based on their choice consistency with human on 2,171 triplets, shows that notable differences remain between LLMs and human. To delve deeper into the reasons behind these differences, we show in Fig. \ref{fig:correlation_matirx}c the most relevant dimensions that humans and models rely on to make choices (see Methods). We see that human and models make different choices because of the differently key dimensions they rely on. For example, human can make choice based on color (like "red"), while LLM only makes choice based on semantics (like "protective"). More examples are in Extended Data Fig. 1f.

	Next, we explored the relationship between the core dimensions of LLMs and humans, as shown in Fig. \ref{fig:correlation_matirx}d. The matrices are generally sparse, indicating that a dimension in one system strongly correlates with only a few dimensions in the other. Many dimensions even show a strong one-to-one mapping. Quantitatively, 31 out of the 66 LLM dimensions and 42 out of the 66 MLLM dimensions strongly correlate with human dimensions ($r>0.4$), indicating substantial alignment. In MLLM, several human dimensions are subdivided (e.g., human dim. 18 "fluid-related" splitting into MLLM dims. 18 "container" and 22 "fluid-related") or amalgamated (e.g., human dims. 3 "animal-related" and 40 "disgusting" merging into MLLM dim. 34 "insect-related"). Similarly, LLM shows adaptations, particularly in semantics, though it lacks sensory dimensions like color or shape. For example, LLM distinguishes between dim. 22 "frozen treats" and dim. 57 "hot drinks" (or dim. 2 "wild animals" vs. dim. 26 "livestock," dim. 13 "vegetables" vs. dim. 18 "fruits," etc.). While MLLM still lacks specific color-related dimensions (e.g., "red," "black"), it aligns more closely with humans, especially in dimensions like shape (e.g., dim. 35 "grainy," dim. 64 "round/curvature") and spatial features (e.g., dim. 8 "serried/stacked," dim. 44 "dense/many small things"). This shows that MLLM, like humans, can perceive a large amount of visual information. Quantitatively, Fig. \ref{fig:correlation_matirx}e shows the number of shared and unique dimensions ($r>0.2$) between models and humans, where 38 of 66 dimensions being shared across the three systems.\\
	
	\begin{figure}[!htbp]
	\centering		
	\includegraphics[width=17cm]{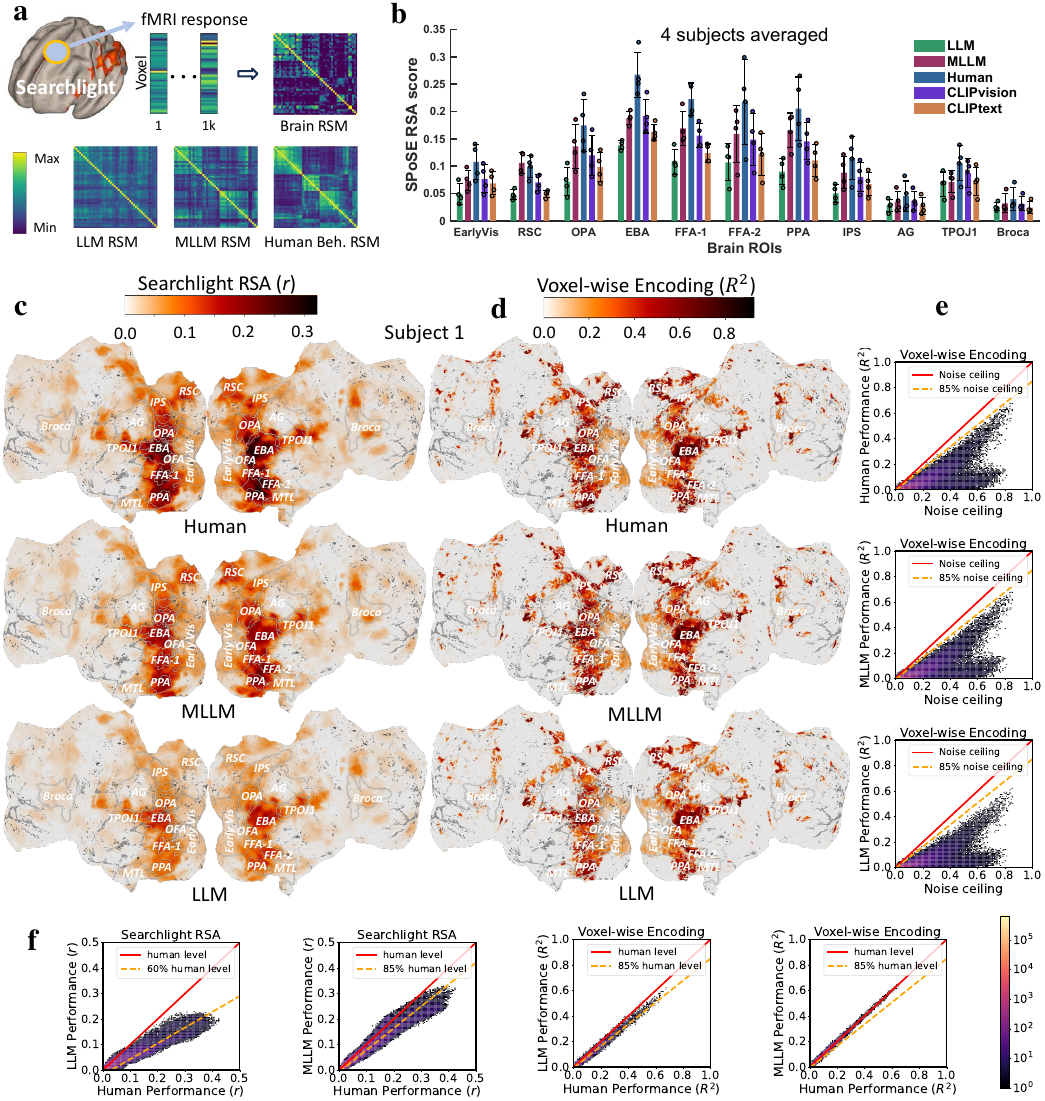}
	\caption{\textbf{Relationship to the cerebral representational geometries.} \textbf{a}, Searchlight brain RSM and the varied model RSMs on the NSD shared\_1k dataset. \textbf{b}, RSA between model RSM and brain ROI RSM constructed from the SPoSE embedding of that brain ROI (see Methods). The error bars reflect SD, and data are presented as mean $\pm$ SD (n=4 subjects, and dots represent the scores of different individuals). \textbf{c-d}, Cortical maps of searchlight RSA and voxel-wise encoding (evaluated by using $R^2$ with noise ceiling normalization). For visualization purpose, we only conducted noise ceiling normalization for voxels that have the predicted $R^2>0.2$. \textbf{e}, 2-D histograms of human, LLM and MLLM performance in $R^2$ against noise ceiling across all voxels in the whole brain. \textbf{f}, 2-D histograms of LLM, MLLM against human performance.}
	\label{fig:rsa}
    \end{figure}

	\noindent \textbf{Relationship to the cerebral representational geometries}
	
	To link LLMs' embeddings with brain responses, we applied searchlight RSA \cite{kriegeskorte2008representational} (see Fig. \ref{fig:rsa}a) using fMRI data from the NSD dataset \cite{allen2022massive}. Independent dimension rating models were fitted for each dimension, and these models predicted multi-dimensional embeddings for objects, creating a representational geometry. We then compared this predicted RSM to SPoSE embedding RSMs of brain ROIs and searchlight RSMs of brain sectors to gauge how well the LLM's embedding aligns with brain regions.

	The representational similarity scores for each model and brain ROI are depicted in Fig. \ref{fig:rsa}b. It should be noted that we adopted the SPoSE method to infer low-dimensional embeddings for CLIP \cite{radford2021learning} (here used as a strong baseline \cite{wang2023better}) and brain ROIs, using cosine distance as a metric to construct the desired odd-one-out records.
	Human and MLLM embeddings outperform LLM and CLIP, particularly in functionally defined, category-selective ROIs (e.g., EBA, PPA, RSC, FFA). However, ROI-based analysis may miss fine-grained spatial patterns, as similar scores can conceal spatial differences.

	Figs. \ref{fig:rsa}c\&d display fine-grained cortical maps of human, LLM, and MLLM embeddings using searchlight RSA and voxel-wise encoding (see Methods) for subject S1, highlighting only significant voxels ($P < 0.05$, FDR-corrected). Additional models and subjects are shown in Extended Data Fig. 8a. Visual inspection shows MLLM and human embeddings align more closely with most of the brain regions than LLM and CLIP, and the contrast of local details can also be clearly viewed. This performance difference is most obvious under searchlight RSA, and relatively moderate in voxel-wise encoding.
	Beyond overall performance metric, peaks in the cortical maps align with scene-selective \cite{epstein2019scene} (PPA, RSC, OPA), body-selective \cite{downing2001cortical} (EBA) and face-selective \cite{sergent1992functional,kanwisher1997fusiform} (FFA, OFA) ROIs, suggesting MLLM captures semantic relationships similar to human cognition. Furthermore, both the overall performance levels and the pattern consistency remain stable across multiple subjects (Extended Data Fig. 8a). Voxel-wise encoding results based on the original CLIP embedding and its low-dimensional SPoSE embedding (Extended Data Fig. 8b) also provide strong evidence that SPoSE is an effective intrinsic dimension learning method.	
	Fig. \ref{fig:rsa}e presents 2-D histograms of human, LLM and MLLM performance in $R^2$ against noise ceiling across all voxels. For human and MLLM, most voxels in the category-selective ROIs (e.g., EBA, PPA, RSC, FFA) are predicted close to their 85\% noise ceiling, while LLM is slightly worse.
	Fig. \ref{fig:rsa}f presents 2D histograms comparing LLM and MLLM to human performance across whole brain voxels. LLM and MLLM achieve about 60\% and 85\% of human performance under searchlight RSA, respectively. In voxel-wise encoding, LLM reaches 90\% of human performance, while MLLM nearly matches human levels.
	
	\section*{Discussion}
	The present study comprehensively investigates object concept representations in LLMs and MLLMs, and their relationship to human cognition and brain representations. We collected 4.7 million behavioral judgments to derive 66 stable dimensions predicting object similarity, uncovering semantic clustering in both LLM and MLLM embeddings, resembling human mental structures. Despite differing architectures, these models developed conceptual representations similar to humans, supported by interpretable dimensions reflecting core aspects of object understanding. MLLM, which integrates visual and linguistic data, predicted individual choices at 85.9\% of the noise ceiling, consistent with findings that multimodal learning enhances representation robustness and generalizability \cite{chang2024survey,minaee2024large,yin2023survey}. Moreover, the strong alignment between MLLM embeddings and neural activity in regions like EBA, PPA, RSC, and FFA suggests that MLLM representations share similarities with human conceptual knowledge \cite{conwell2022can}.\\
	
	\noindent \textbf{Broad applications of the derived embeddings}
	
	The low-dimensional mental embeddings identified in this study can be used in human-machine representation alignment and fusion, potentially enhancing human-machine interfaces and collaborative systems by revealing shared object representation schemas. Practically, these interpretable dimensions could inform the development of more human-like artificial cognitive systems, improving their natural interaction with humans \cite{zador2023catalyzing}.  To better align LLM and MLLM with human reasoning in the odd-one-out task, we can explore the method of guiding model attention to human-preferred dimensions. By tailoring prompts to emphasize specific attributes (e.g., "red" or "artificial"), we believe that models could make choices more consistent with human judgments (i.e., explicit guidance can help bridge the gap between model and human reasoning; Supplementary Figs. 2-4).
	Moreover, the collected extensive machine behavioral datasets offer a valuable benchmark for evaluating AI model representations.\\

	\noindent \textbf{Relationship to the other related studies}
	
	Both the human brain and large-scale AI models are complex systems, typically analyzed through dimensionality reduction. Recent hypotheses like the "low-rank" \cite{thibeault2024low} and "distributed information bottleneck" \cite{murphy2024information} propose solutions to identifying optimal latent dimensions. Our findings align with these concepts, demonstrating that LLMs can develop human-like object representations using fundamental dimensions, akin to the brain's capacity to derive rich conceptual knowledge from simple neural mechanisms. Exploring these low-dimensional structures could deepen our understanding of cognition in both biological and artificial systems.
	
	The similarity between LLMs and human representations, despite differing input modalities, suggests a convergence beyond data covariance. This is consistent with findings on innate semantic transformations in the visual system \cite{doerig2022semantic}, and is further supported by the interpretability of LLMs' embeddings, reflecting fundamental semantic structures. Prior studies \cite{conwell2023ccn,mcmahon2024language,conwell2024monkey} demonstrate that artificial models can predict visual brain activity, which aligns with our results showing model-neural correlations in higher cortical regions. These findings suggest LLMs develop representations that capture key aspects of human conceptual knowledge \cite{tuckute2024language,tuckute2024driving}, further highlighting the natural alignment between language and vision \cite{popham2021visual,roads2020learning}.
	Previous fMRI studies have revealed diverse organizational principles in the brain for processing external stimuli. The primary visual cortex exhibits retinotopy through eccentricity and angle selectivity \cite{sereno1995borders,engel1997retinotopic}. These principles of dimensional organization extend to higher-order information \cite{hansen2007topographic,huth2012continuous,harvey2013topographic,sha2015animacy,huth2016natural,margulies2016situating,huntenburg2018large}. Our study expands this research to the conceptual representations of natural objects.
	
	Traditionally, neural network representations are analyzed by examining neuron activation patterns \cite{bau2020understanding,mcgrath2022acquisition,achtibat2023attribution,bills2023language}. However, as AI systems grow in complexity, neuron-level approaches become less effective. Instead, inspired by cognitive psychology, behavioral methods can infer AI system representations through actions. Decades of research have developed techniques to elucidate mental representations from human behavior \cite{sanborn2010uncovering,hebart2020revealing}. Our study adopts this behavioral approach for LLMs, complementing existing neuron-level methods.
	Probing LLMs from a cognitive perspective has gained attention \cite{mahowald2024dissociating,qu2024integration,demszky2023using,meng2024ai,marjieh2023language,campbell2024human}, revealing insights into areas like color processing \cite{kawakita2023comparing}, emotion analysis \cite{li2023large,sabour2024emobench}, memory \cite{janik2023aspects,huff2024towards}, morality \cite{schramowski2022large}, and decision-making \cite{binz2023using,peterson2021using,alsagheer2024comparing}. Understanding the parallels between human cognition and LLMs offers exciting opportunities to explore the intersections of AI and cognitive science \cite{messeri2024artificial,zador2023catalyzing}.\\
	
	\noindent \textbf{Limitations and future directions}
	
	One potential limitation of this study is its focus on ChatGPT-3.5 and Gemini Pro Vision (v1.0), which may not encompass the full spectrum of models. However, the methodology is extendable to other state-of-the-art LLMs such as GPT-4V \cite{achiam2023gpt}. This extension could reveal the generalization of identified dimensions and highlight the unique aspects of different AI architectures. Another potential limitation is that the impact of varying language prompts on LLMs' responses. In this study, the language prompts we used were carefully designed to ensure that the LLMs understand the task instructions correctly. We think that these considerations have a negligible impact on the study's overall conclusions. Moreover, we only employed object-level annotations in the language prompts of LLM. Object-level annotations focus on abstract categories, while image-level annotations (generated by a vision-language model or human annotators) can capture more image-specific visual attributes like color and texture (Supplementary Fig. 5). Using the image-level annotations will make LLM more consistent with human judgments (this can be confirmed in the MLLM probing experiments, which is equivalent to using image-level annotation in essence), highlighting the importance of visual information in similarity judgments (Supplementary Figs. 6-8).
	
	Future work could leverage instruction fine-tuning for LLM/MLLM on large-scale triplet odd-one-out question-answer pairs, where answers include both human choices and the underlying reasoning dimensions, to improve model-human alignment.\\
	
	\section*{Methods}
	
	\noindent \textbf{Stimuli and triplet odd-one-out task.} 
	In selecting stimulus objects, our preference was for the THINGS database \cite{hebart2019things}, a resource designed to encompass 1,854 living and non-living objects based on their practical usage in daily life. During the triplet odd-one-out task, participants (humans or LLMs) encountered three objects drawn from the THINGS database, either through images or textual descriptions. Their objective was to identify the object with the highest dissimilarity among the three. This task evaluates the relationship between two objects considering the context set by a third object. Featuring a diverse range of objects, this method provides a systematic means to assess perceived similarity unaffected by context, thus minimizing response bias. Moreover, it enables the measurement of context-dependent similarity, such as by restricting similarity evaluations to specific higher-level categories like animals or vehicles.\\
	
	\noindent \textbf{Behavioral responses from humans.} 
	The human behavioral dataset utilized in our research originated from a recent study \cite{hebart2023things}, where 5,517,400 human similarity judgments were collected via Amazon Mechanical Turk. After quality control--which excluded 818,240 trials (14.83\%) based on overly fast responses (\textgreater25\% trials \textless800ms and \textgreater50\% \textless1,100ms), repetitive patterns (outside central 95\% distribution in $\geq$200 trials), and inconsistent demographic reporting (\textgreater3 ages provided)--the final dataset comprised 4,699,160 valid trials from 12,340 participants. Participants (6,619 female; 4,400 male; 56 other/unspecified; mean age = 36.71 years, SD = 11.87; 41.9\% unreported age) were right-handed with normal/corrected vision, compensated at \$0.10 per 20 trials. The protocol, approved by the NIH Institutional Review Board (93-M-0170) and NIH Office of Human Research Subject Protection, obtained informed consent. While self-selection bias (tech-savvy English-speakers) and handedness exclusion may limit generalizability, the focus on relative similarity judgments--demonstrated robust across demographics \cite{hebart2020revealing}--reduces population-specific effects.\\

	\noindent \textbf{Collecting behavioral responses from LLM.}  
	For our study, we gathered all human-used similarity judgments, totaling 4.7 million trials. To solicit responses from ChatGPT-3.5 (gpt-3.5-turbo),  Llama3.1 (Meta-Llama-3.1-8B-Instruct), and GPT-4 (gpt-4-0314), we employed a prompt where each image was represented by its object name and descriptions, as image input processing was not supported by these models. These text descriptions are sourced from definitions of object names in WordNet, Google, or Wikipedia, and have been compiled and made publicly available at \url{https://osf.io/jum2f/}. For model comparison, Llama3.1 was used to collect the full sampling of triplets (91,568 trials) of the 48 typical objects. Due to cost constraints, GPT-4 only amassed a total of 2,171 trials, primarily for initial comparisons with ChatGPT-3.5.
	
	The prompt structure used was standardized:
	"\emph{Given a triplet of objects \{[Object\_A], [Object\_B], [Object\_C]\}, which one in the triplet is the odd-one-out? Please give the answer first and then explain in detail.}" In practice, [Object\_A], [Object\_B], and [Object\_C] were replaced with the respective object descriptions for each trial. The temperature parameter, dictating response randomness in LLMs, was set to 0.01. 
	Because of the well-structured nature of the model's responses, we parsed the model choice from the first sentence of their response using string matching.
	To assess the upper limit of predictability under dataset randomness (the noise ceiling), we randomly selected 1,000 triplets and conducted a minimum of 14 trials and a maximum of 25 trials for each using the same prompt, evaluating consistency in choices across trials. \\
	
	\noindent \textbf{Collecting behavioral responses from MLLM.} Regarding collecting behavioral responses from Gemini Pro Vision (v1.0), we adopted a similar strategy. The prompt we used is as follows:
	"\emph{You are shown three object images side by side and are asked to report the image that was the least similar to the other two. You should focus your judgment on the object, but you are not given additional constraints as to the strategy you should use. If you did not recognize the object, you should base your judgment on your best guess of what the object could be. 1. Tell me your answer. 2. Tell me the location of the object you have chosen. 3. Explain the reasons.}"  In some trials, the Gemini Pro Vision model refused to respond because it believed that the given images contained some unknown sensitive information. In this case, we applied a method akin to image replacement to address the issue.
	
	The temperature parameter for determining response randomness in Gemini Pro Vision was also configured to 0.01, with images displayed at 512 x 512 pixels. Since the model's responses are well structured, we extracted the keyword about the position of the object in its answers (e.g., "left," "middle," or "right") to determine the model's choice. Similarly, to gauge the noise ceiling and potential predictability, we additionally sampled 1,000 randomly chosen triplets and ran a minimum of 14 trials and a maximum of 25 trials for each of them using the same prompt for each trial and estimated the consistency of choices for each triplet across trials. 
	
	As for the model of Qwen2\_VL-7B, we used a similar strategy to collect the full sampling of triplets for the 48 typical objects.
	\\

	\noindent \textbf{Constructing behavioral responses for the other models.} 	For models do not have visual or language-based question-answer capabilities (such as CLIP, SimCLR, VGG16, etc.), we first used the pre-trained model to extract the features of the object images (or their language descriptions), and then constructed the required odd-one-out data based on the cosine distance of the features. \\
	
	\noindent \textbf{Feature extractors.} 	For the pre-trained models originally used for classification tasks (such as VGG16, ResNet18, etc.), we extracted the penultimate layer features, rather than the head. For CLIP, we extract features in the final embedding layer. 
	For GPT2 and Llama3.1, we extracted features by averaging the last hidden state activations across all tokens to obtain sentence embeddings. For Qwen2\_VL, we extracted image features from the last layer of its visual branch, which is based on a 600M-parameter ViT.
	Some of the pretrained models sourced from the following repositories: the Torchvision model zoo, the Pytorch-Image-Models (timm) library, the VISSL (self-supervised) model zoo, the OpenAI CLIP collection, and the Transformer python library.
	In particular, the Gabor model feature extractor consists of a single fixed set of convolutions: 12 Gabor wavelets with spatial frequency log-spaced between 3 and 72 cyc/stimulus at 6 evenly-spaced orientations between 0 and $\pi$, following previous work \cite{st2023brain}. \\

	\noindent \textbf{Natural Scene Dataset (NSD).}
	NSD \cite{allen2022massive}, recognized as the largest neuroimaging dataset linking brain insights with artificial intelligence, involves richly sampled fMRI data from 8 subjects. Across 30-40 MRI sessions, each subject observed between 9,000-10,000 distinct natural scenes using whole-brain gradient-echo EPI at 1.8 mm isotropic resolution and 1.6 s TR during 7T scanning. Image stimuli were drawn from the COCO dataset \cite{lin2014microsoft}, with corresponding captions retrievable using COCO ID. To assess the generalization ability of the low-dimensional embeddings learned from humans and LLMs across datasets, the shared\_1k subset from the NSD were chosen as the test set (because the stimuli in this subset were shared by all 8 subjects). Additionally, fMRI responses linked to the shared\_1k stimuli across subjects S1, S2, S5, and S7 were earmarked for subsequent analysis (because subjects S3, S4, S6, and S8 did not complete the full fMRI data acquisition).\\
	
	\noindent \textbf{Sparse Positive Similarity Embedding (SPoSE).}
	Utilizing the SPoSE approach \cite{zheng2019revealing,hebart2020revealing}, we derived embedding representations for 1,854 objects based on similarity judgment data from LLM and MLLM, respectively. The PyTorch implementation for this process can be accessed at \url{https://github.com/ViCCo-Group/SPoSE}. Initially, an embedding matrix $\mathbf{X}$ was created with random weights in the range of 0 to 1 across 100 latent dimensions for each object, resulting in a 1854-by-100 matrix. Stochastic gradient descent was subsequently applied to fine-tune this embedding matrix using odd-one-out responses. The optimization objective function aimed to minimize a combination of cross-entropy loss concerning triplet choice probabilities for all options and an L1-norm on the weights to promote sparsity:
	\begin{align}
		\hspace{2.3cm} \min \mathcal{L}(\mathbf{x}) = \sum^n \log \left(\frac{\exp \left(\mathbf{x}_i \mathbf{x}_j\right)}{\exp \left(\mathbf{x}_i \mathbf{x}_j\right)+\exp \left(\mathbf{x}_i \mathbf{x}_k\right)+\exp \left(\mathbf{x}_j \mathbf{x}_k\right)}\right)+\lambda \sum^m\|\mathbf{x}\|_1,
	\end{align}
	where $\mathbf{x}$ corresponds to an object vector; $i$, $j$ and $k$ to the indices of the current triplet; $n$ to the number of triplets; and $m$ to the number of objects. The regularization parameter $\lambda$, which controls the trade-off between sparsity and model performance, was determined using cross-validation on the training set ($\lambda = 0.004$ for LLM, $0.0035$ for MLLM, $0.00385$ for humans, and 0.007 for the other models and brain ROIs). In addition to sparsity, the optimization was constrained by strictly enforcing weights in the embedding $\mathbf{X}$ to be positive. The minimization of this objective was carried out using stochastic gradient descent with an Adam optimizer \cite{kingma2014adam} (with default parameters) and a batch size of 100 on triplet odd-one-out judgments. After the optimization was complete, dimensions with weights below 0.1 for all objects were eliminated. Finally, the dimensions underwent sorting based on the sum of their weights across objects in descending order.

	This model operates under two key theoretical assumptions. Firstly, it postulates sparsity within the embedding space dimensions, indicating that each object primarily influences certain dimensions rather than all. Secondly, it assumes positivity in these dimensions. Consequently, an object's weight on a specific dimension signifies the extent of the related property within the object. These assumptions diverge from typical dimensionality reduction approaches like Principal Component Analysis (PCA), which assume dense dimensions across the real number spectrum. Furthermore, SPoSE facilitates cross-correlations among dimensions while PCA assumes independence. Consequently, SPoSE often uncovers a greater number of dimensions, reflecting finer details or attributes, which are more easily interpretable compared to PCA dimensions. Notably, the weight an object holds on a dimension directly corresponds to the presence of the associated property within the object.
	
	We opted for the behavioral odd-one-out task and the SPoSE method to learn the low-dimensional embeddings of LLMs rather than attempting to directly access their internal features, primarily due to the challenges associated with extracting features from modern, large-scale LLMs that are often proprietary or too vast to navigate directly. This approach allows us to circumvent the limitations imposed by the closed nature or sheer scale of contemporary LLMs, providing us with a more feasible avenue to explore their mental representations.
	\\
	
	\noindent \textbf{Reproducibility of embedding dimensions.}
	Considering the stochastic nature of the optimization process, the SPoSE method yields varying sets of dimensions upon each reiteration. To assess the stability of the 66-dimensional embedding, we conducted 20 model runs with distinct random initializations. Evaluating each original dimension against all dimensions in the 20 reference embeddings, we identified the best-matching dimension based on the highest correlation. Consistent with previous research \cite{hebart2020revealing}, a Fisher z-transform was applied to these correlations, averaged across the 20 reference embeddings, and then reversed to obtain a mean reliability value for each dimension across all 20 embeddings.\\
	
	\noindent \textbf{Category prediction.}
	Evaluating the representational embeddings' categorization performance involved testing them across 18 out of the 27 THINGS database categories. Objects falling into multiple categories were excluded from the analysis, resulting in the removal of 9 categories. Among these excluded categories, 7 were subcategories or had less than ten unique objects post-filtering. The remaining 18 categories included clothing, toy, vehicle, container, electronic device, animal, furniture, body part, food,  musical instrument, plant, home decor, sports equipment, office supply, part of car, medical equipment, tool, and weapon, totaling 1,112 objects. Classification was conducted through leave-one-object-out cross-validation. Training involved computing category centroids by averaging the 66-dimensional vectors of all objects within each category, excluding the left-out object. The category membership of the excluded object was predicted based on the smallest Euclidean distance to the respective centroid. This process was iterated for all 1,112 objects, with prediction accuracy averaged across the dataset. The chance level is determined by 1000 permutation tests.\\
	
	\noindent \textbf{Evaluating consistency between humans and models by comparing behaviors.}
	With the exception of GPT-4, all other models (and human) have completed behavioral data acquisition on the full sample triples of the 48 typical objects described above.  For each model, we constructed its RSM for the 48 objects by calculating the choice probability of each object pair.  To estimate human consistency, following previous work \cite{rajalingham2018large}, we computed the Pearson correlation on the behavioral RSMs from the model ($m$) and the human ($h$) and we then divide that raw Pearson correlation by the geometric mean of the split-half internal reliability measured for each system as follows:
	\begin{align}
		\hspace{3cm} \tilde{\rho}(m, h) = \frac{\rho(RSM_m, RSM_h)}{\sqrt{\rho(RSM_{m}^{half_1}, RSM_{m}^{half_2}) \rho(RSM_{h}^{half_1}, RSM_{h}^{half_2})}},
	\end{align}
	where $RSM_{m}^{half_1}$ and $RSM_{m}^{half_2}$ were computed by using the split-half behavioral data of triples of the 48 typical objects, and similar for $RSM_{h}^{half_1}$ and $RSM_{h}^{half_2}$.
	
	Comparison between ChatGPT-3.5 and GPT-4 was conducted directly based on their choice consistency with human on a specific set of 2,171 triplets. We conducted a total of 5 comparisons, each based on randomly selecting 1,000 samples from these 2,171 samples, and finally reported the average result. \\
	
	\noindent \textbf{Dimensional relevance score for odd-one-out choice.}
	For a given triplet, we compute the original predicted softmax probabilities based on the entire low-dimensional embeddings of each image within the triplet. Then, we iteratively remove a certain dimension from the low-dimensional embeddings, calculate the softmax probabilities predicted by the pruned embeddings, and then compute the difference between the softmax probabilities obtained before and after pruning. This difference is taken as the relevance score for that dimension. This approach has been used in a previous study \cite{mahner2024dimensions}. \\

	\noindent \textbf{Dimension naming.}
	In defining the human mental embedding, the dimension names from a previous investigation were employed as references \cite{hebart2023things}. However, for LLM and MLLM, each of the 66 dimensions within the embedding was associated with common-sense labels through a straightforward naming procedure. Specifically, we analyze a set of 1-by-12 images of objects and identify shared properties described in the images. Each array consisted of images selected from the top of one dimension from the embedding. Ten of the authors provided concise labels, limited to 1–2 words, describing the arrayed images. Subsequently, word clouds were generated to visualize dimension names, showcasing the distribution of labels based on frequency, utilizing the wordcloud function in MATLAB (Mathworks) with default settings. Finally, the lead authors of this study gave intuitive labels for each dimension.  Dimension labels were also summed up by the MLLM (here gemini-pro-1.5-exp) with the prompt as follows: "\emph{There are 9 subfigures in the picture. Please use 1-2 English words or phrases to describe the common theme represented by these 9 subfigures.}"\\

	\noindent \textbf{Dimension rating for NSD images.}
	We predicted the 66 object dimensions for each image within the NSD dataset. Specifically, we leveraged the OpenAI-trained CLIP model \cite{radford2021learning} (with "ViT-L/14" as the backbone), which is a multimodal model trained on image-text pairs and which was recently demonstrated to yield excellent prediction of human similarity judgments \cite{hebart2022efficiently,muttenthaler2022human}. For each of the 1,854 object images in the THINGS dataset, we extracted the image and text features from the final layer of the CLIP image and text encoders, respectively. Subsequently, for each of the 66 dimensions of LLM (or MLLM, or Human), we fitted a ridge regression model to predict dimension values, using a concatenation of the extracted image and text features from CLIP as input. The optimal regularization hyperparameters were determined by using 5-fold cross-validation across the training set (100 candidate parameters spaced evenly on a log scale from $10^{-3}$ to $10^{3}$, that is np.logspace(-3, 3, 100)) . These trained regression models were then applied to the  extracted features across all images in the NSD dataset. 
	\\
	
	\noindent \textbf{Searchlight RSA.}
	For fMRI, local cerebral RSMs were computed in subject space within a grey-matter spherical region (6 mm diameter) centered at each voxel location.  RSA analyses assessed the Pearson correlation $r$ between the local cerebral RSM and each kind of the model RSMs.\\
	
	\noindent \textbf{SPoSE RSA.}
	For each brain ROI, we extracted the fMRI signal in that region on the shared\_1k dataset and constructed a large number of odd-one-out data based on the cosine distance. After that, SPoSE learning was used to obtain the corresponding low-dimensional embeddings of each brain ROI, and the RSMs of each ROI were calculated using the learned low-dimensional embeddings. Finally, Pearson correlations between the brain ROI RSM and the model RSM were calculated.\\
	
	\noindent \textbf{Voxel-wise encoding.}
	For each subject in the NSD, we built a ridge regression model to predict the fMRI response to each test image per voxel.  The images of the training set are subject-specific, but the images of the test set are shared (that is, shared\_1k). For all training and testing images, we first used the dimension rating model to predict the low-dimensional embeddings, and then conducted voxel-wise fitting based on the predicted embeddings. The regularization parameter for each voxel was selected autonomously through a 5-fold cross-validation process on the training dataset. We explored 100 evenly spaced regularization parameters on a logarithmic scale ranging from $10^{-3}$ to $10^{3}$, which corresponds to the np.logspace(-3, 3, 100) function in Python. The model's accuracy was assessed on the test dataset utilizing both Pearson's correlation coefficient ($r$) and the noise ceiling normalized coefficient of determination ($R^2$). Following the NSD work \cite{allen2022massive}, the noise ceiling was calculated by:
	\begin{align}
		\hspace{6cm}	NC = 100 \times \frac{ncsnr^2}{ncsnr^2 + \frac{1}{n}},
	\end{align}
	where $n$ indicates the number of trials that are averaged together ($n=3$ for subjects S1, S2, S5, and S7), and $ncsnr$ indicates the noise ceiling signal-to-noise ratio which has been provided in NSD.
	To ascertain the statistical significance of our predictions, we conducted a bootstrapping procedure, resampling the test dataset with replacement 2,000 times, and subsequently calculated the False Discovery Rate (FDR) adjusted $P$-values.\\

	\noindent \textbf{Abbreviation of Brain ROIs.}
	EarlyVis: early visual cortex; Scene, PPA: parahippocampal place area, OPA: occipital place area, RSC: retrosplenial cortex; Body, EBA: extrastriate body area; Face, FFA-1: fusiform face area 1, FFA-2: fusiform face area 2; Mind and Language, TPOJ-1: temporoparietal junction 1, AG: angular gyrus, Broca, MTL: medial temporal lobe.\\
	
	\noindent \textbf{Visualization of cerebral cortex.}
	To visualize the analytical outcomes across the entire cortical region, we employed flattened cortical surfaces derived from individual subjects' anatomical images. FreeSurfer \cite{fischl2012freesurfer} facilitated the generation of cortical surface meshes from T1-weighted anatomical images. This process involved applying five relaxation cuts on each hemisphere's surface and excluding the corpus callosum. Subsequently, functional images were registered to the anatomical images and mapped onto the surfaces for visualization purposes using Pycortex \cite{gao2015pycortex}.
	\\

	\section*{Data availability}
	
	The THINGS database is accessible at \url{https://osf.io/jum2f/}. The behavioral triplet odd-one-out datasets for Human, ChatGPT-3.5, and Gemini Pro Vision 1.0 can be found at \url{https://osf.io/f5rn6/}, \url{https://osf.io/qn5uv/}, and \url{https://osf.io/qn5uv/}, respectively. Those interested in the preprocessed NSD fMRI dataset supporting this research can obtain it from \url{http://naturalscenesdataset.org/}. Language descriptions for the 1,854 THINGS objects, the learned mental embeddings of LLM and MLLM, as well as the human and MLLM annotated dimension names are shared in \url{https://osf.io/qn5uv/}.

	\section*{Code availability}
	
	The code used for data collection, embedding learning, dimension rating, result analysis, and visualization in this study is publicly available on GitHub (\url{https://github.com/ChangdeDu/LLMs_core_dimensions}\cite{du2025llm}).

	\section*{Acknowledgements} 
	
	This work was supported in part by the Strategic Priority Research Program of the Chinese Academy of Sciences (Grant No. XDB1010202); in part by the National Natural Science Foundation of China under Grant 62020106015 and Grant 62206284; in part by Beijing Natural Science Foundation under Grant L243016, and in part by the Beijing Nova Program under Grant 20230484460. We would like to thank Martin N. Hebart for sharing the THINGS database and 4.7 million human behavioral responses. We also thank Emily J. Allen and Kendrick Kay for sharing the NSD fMRI data. All illustrative images in this article were sourced from Pixabay and Pexels due to copyright restrictions.
	
	\section*{Author contributions}
	
	C.D. and H.H. designed the research. C.D. conducted the experiments. C.D., Y.S, K.F., and J.P. collected the data. C.D. wrote the paper.  C.D., B.W., W.W., Y.G., S.W., C.Z., J.L., S.Q., L.C. and H.H. analyzed the results.  All authors read and approved the paper.
	
	\section*{Competing interests} 
	
	The authors declare no competing interests.

\newpage
	\section*{Extended data}

\beginextended
\extendeddatafigure	
\extendeddatatable	

\begin{figure}[!htbp]
	\centering
	\includegraphics[width=17cm]{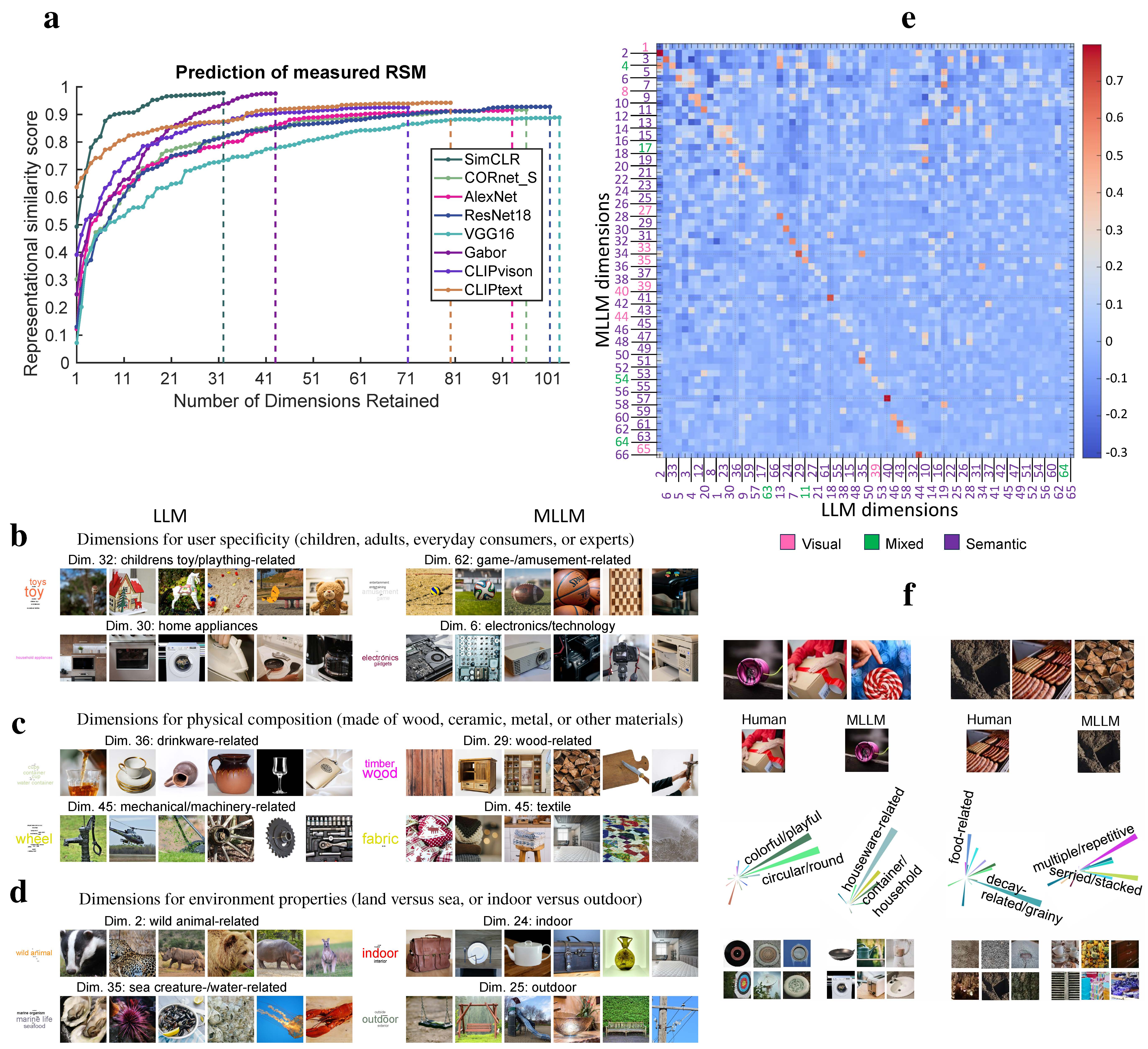}
	\caption{\textbf{Object dimensions learned by different models and their interpretations(related to Figs. \ref{fig:66dimensional}, \ref{fig:dim_vis} and \ref{fig:correlation_matirx}).} \textbf{a}, Dimensions retained by different models and the ability to predict their behavioral RSMs. \textbf{b-d}, Object dimensions illustrating their interpretability for LLM and MLLM. \textbf{e}, Cross-correlation matrix between LLM and MLLM. \textbf{f}, Key dimensions that underpin the different choices that humans and models made.}
\label{fig:dim_vis_2}
\end{figure}

\begin{figure}[!htbp]
	\centering
    \includegraphics[width=17cm]{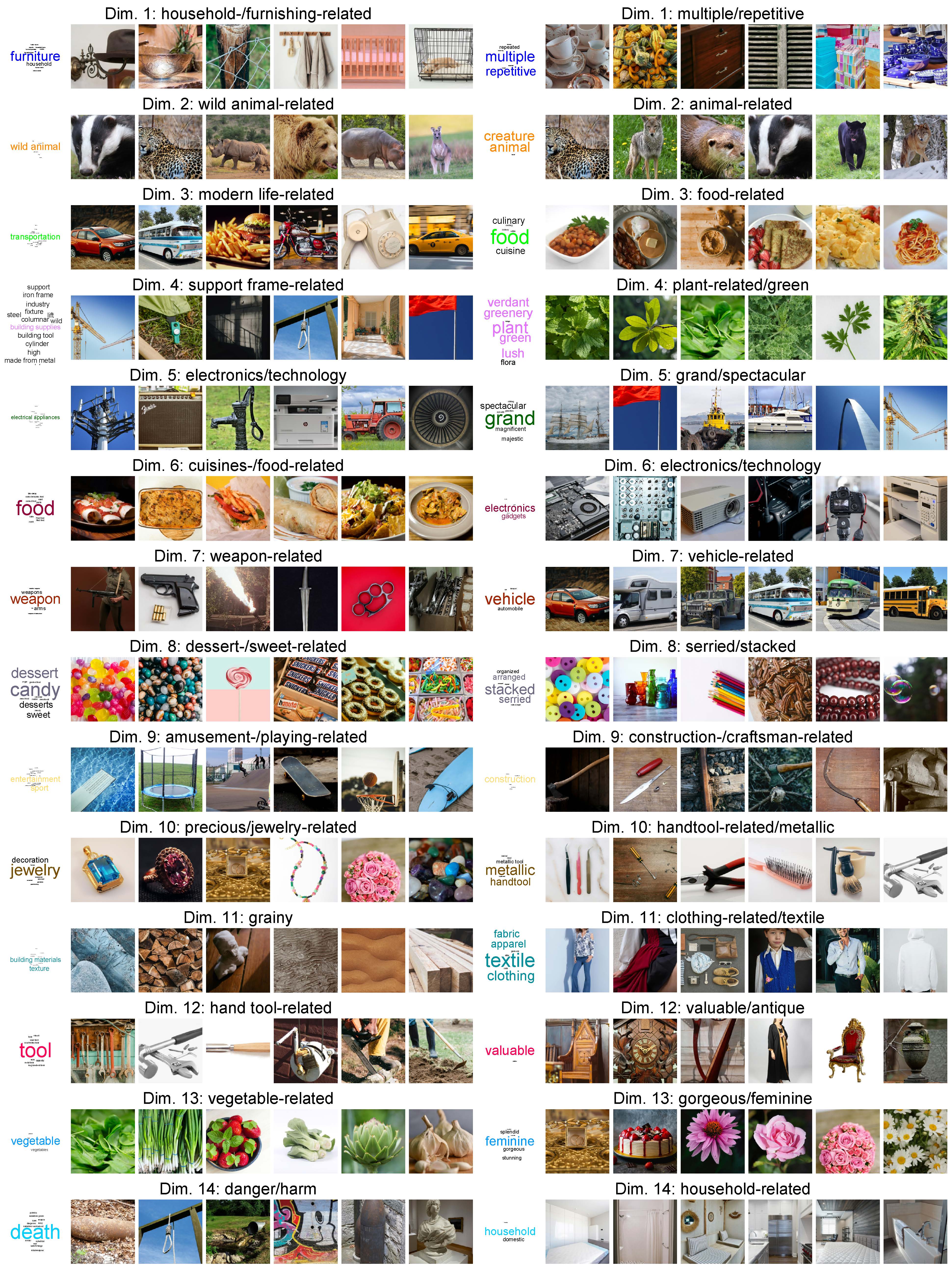}
	\caption{\textbf{Object dimensions (1-14) illustrating their interpretability for LLM (left) and MLLM (right)(related to Fig. \ref{fig:dim_vis}).} Each dimension is illustrated with the top 6 images with the highest weights along this dimension.}
	\label{fig:dim_vis_1_14}
\end{figure}

\begin{figure}[!htbp]
	\centering	
    \includegraphics[width=17cm]{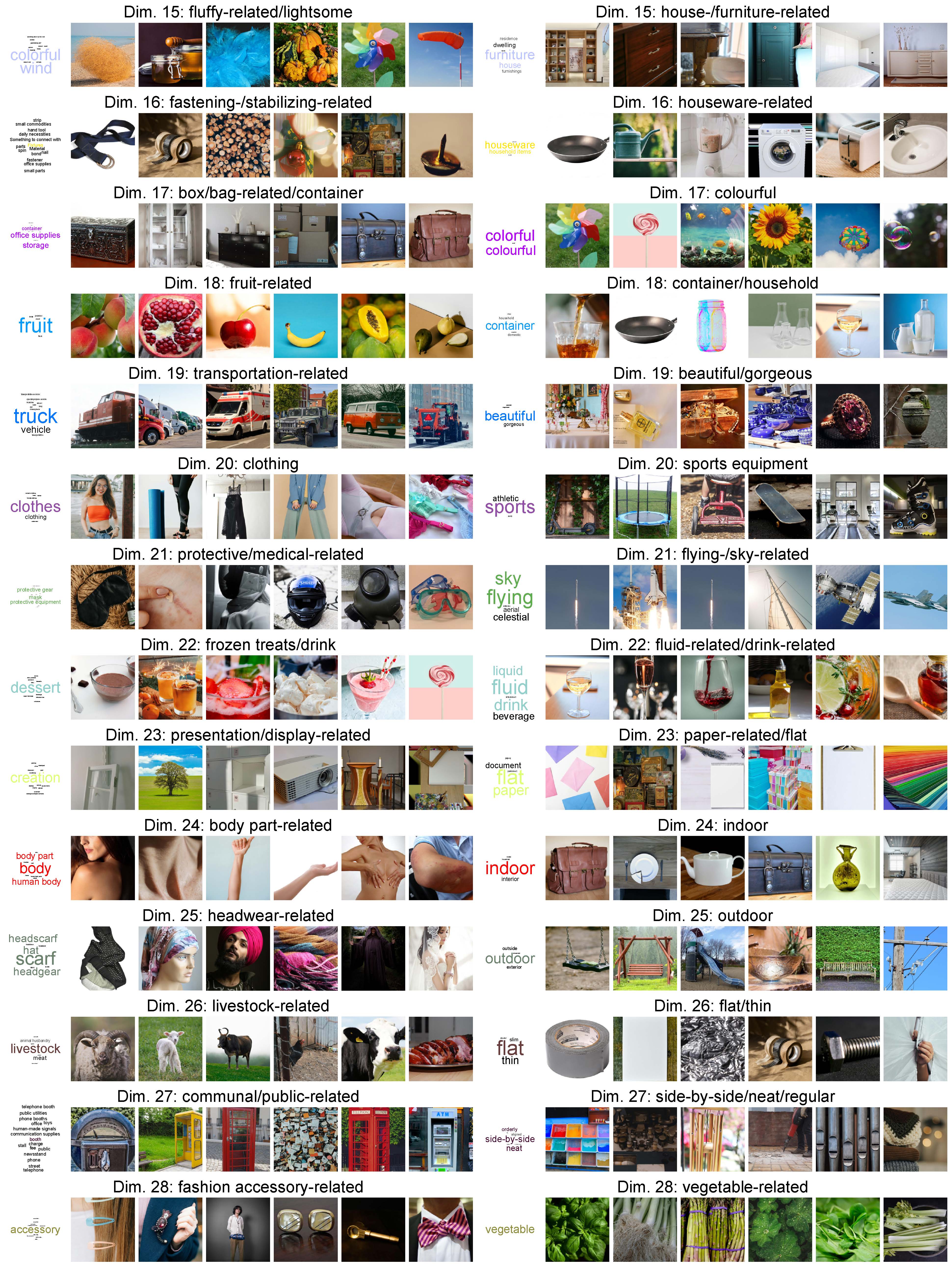}
	\caption{\textbf{Object dimensions (15-28) illustrating their interpretability for LLM (left) and MLLM (right)(related to Fig. \ref{fig:dim_vis}).} Each dimension is illustrated with the top 6 images with the highest weights along this dimension.}
	\label{fig:dim_vis_15_28}
\end{figure}

\begin{figure}[!htbp]
	\centering
    \includegraphics[width=17cm]{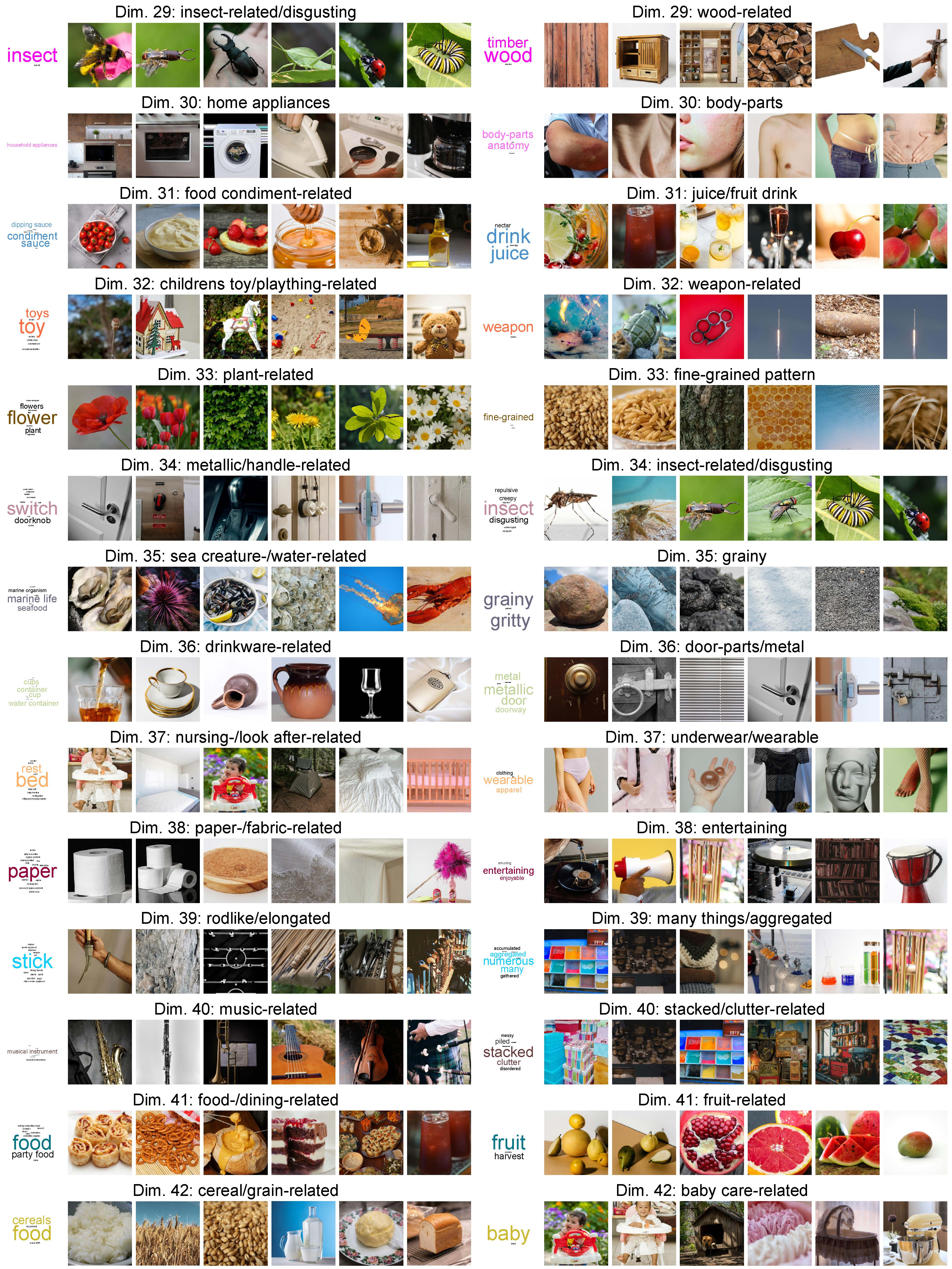}
	\caption{\textbf{Object dimensions (29-42) illustrating their interpretability for LLM (left) and MLLM (right)(related to Fig. \ref{fig:dim_vis}).} Each dimension is illustrated with the top 6 images with the highest weights along this dimension.}
	\label{fig:dim_vis_29_42}
\end{figure}

\begin{figure}[!htbp]
	\centering
    \includegraphics[width=17cm]{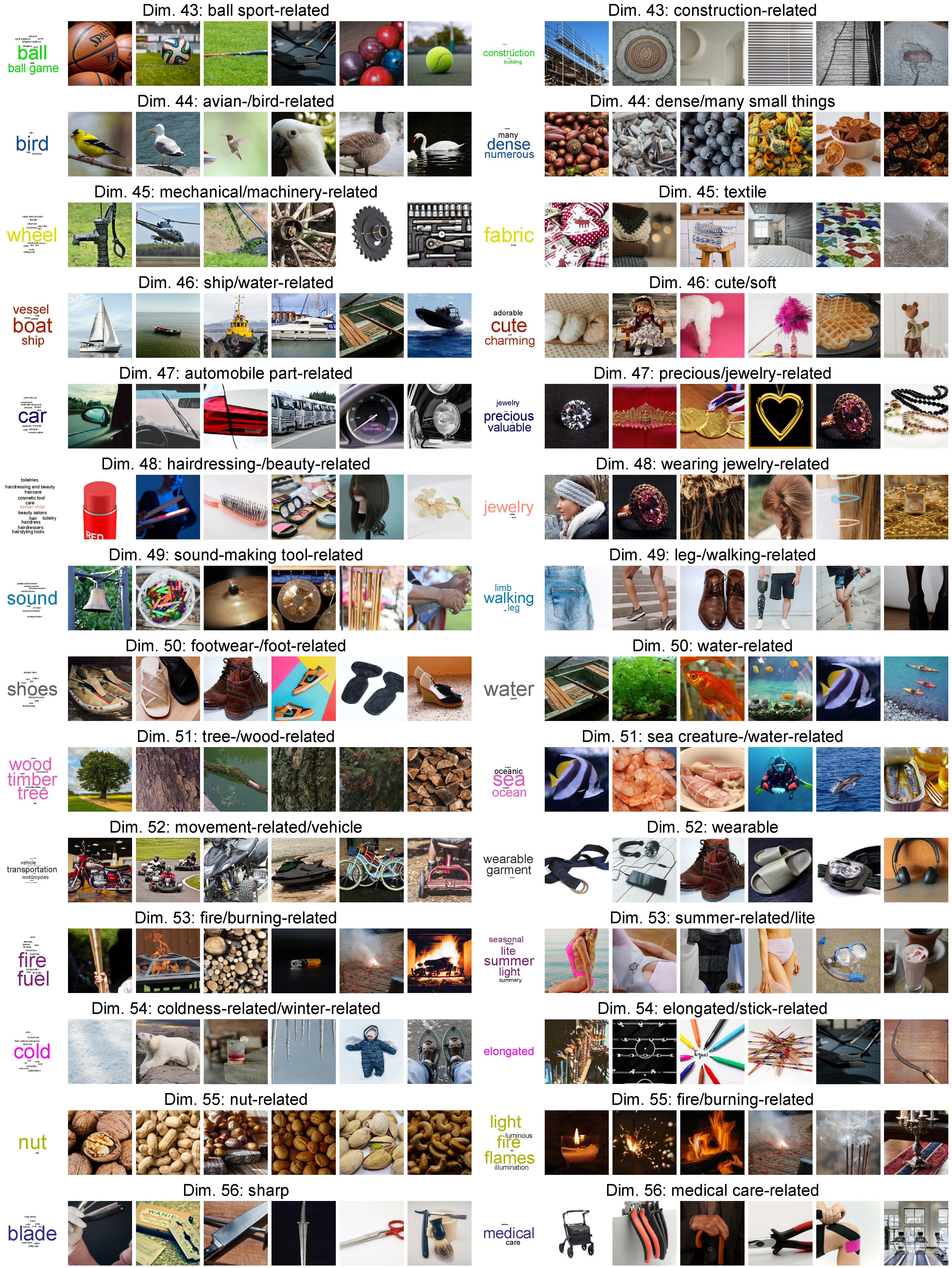}
	\caption{\textbf{Object dimensions (43-56) illustrating their interpretability for LLM (left) and MLLM (right)(related to Fig. \ref{fig:dim_vis}).} Each dimension is illustrated with the top 6 images with the highest weights along this dimension.}
	\label{fig:dim_vis_43_56}
\end{figure}

\begin{figure}[!htbp]
	\centering
    \includegraphics[width=17cm]{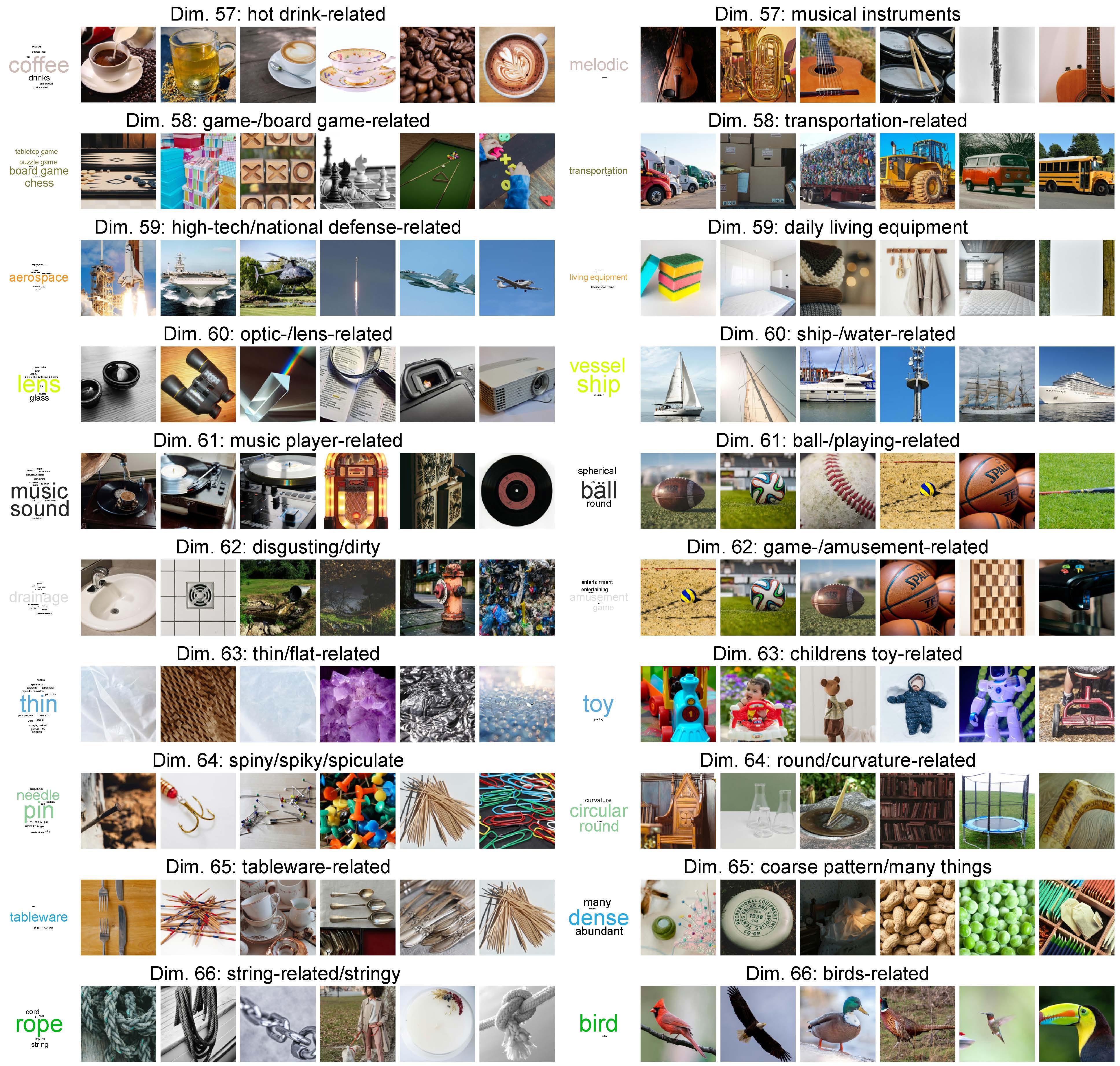}
	\caption{\textbf{Object dimensions (57-66) illustrating their interpretability for LLM (left) and MLLM (right)(related to Fig. \ref{fig:dim_vis}).} Each dimension is illustrated with the top 6 images with the highest weights along this dimension.}
	\label{fig:dim_vis_57_66}
\end{figure}

\begin{table}[!htbp]
	\centering
	\ra{0.9}
	\co{1.5pt}
	\caption{List of all dimensions and their intuitive labels summed up by the human experts (related to Fig. \ref{fig:dim_vis}).}
	\resizebox{17cm}{!}{
		\begin{tabular}{|c|p{6cm}|p{6cm}|p{6cm}|}
			\hline
			\textbf{Dim. No.} & \textbf{\qquad\qquad LLM (GPT3.5-Turbo)} & \textbf{\qquad MLLM (Gemini Pro Vision 1.0)} & \textbf{\qquad\qquad\qquad Humans} \\
			\hline
			1 & household-/furnishing-related & multiple/repetitive & metallic/artificial \\ \hline
			2 & wild animal-related & animal-related & food-related \\ \hline
			3 & modern life-related & food-related & animal-related \\ \hline
			4 & support frame-related & plant-related/green & textile \\ \hline
			5 & electronics/technology & grand/spectacular & plant-related \\ \hline
			6 & cuisines-/food-related & electronics/technology & house-related/furnishing-related \\ \hline
			7 & weapon-related & vehicle-related & valuable/precious \\ \hline
			8 & dessert-/sweet-related & serried/stacked & transportation-/movement-related \\ \hline
			9 & amusement-/playing-related & construction-/craftsman-related & body-/people-related \\ \hline
			10 & precious/jewelry-related & handtool-related/metallic & wood-related/brown \\ \hline
			11 & grainy & clothing-related/textile & electronics/technology \\ \hline
			12 & hand tool-related & valuable/antique & colorful/playful \\ \hline
			13 & vegetable-related & gorgeous/feminine & outdoors \\ \hline
			14 & danger/harm & household-related & circular/round \\ \hline
			15 & fluffy-related/lightsome & house-/furniture-related & paper-related/flat \\ \hline
			16 & fastening-/stabilizing-related & houseware-related & sports-/playing-related \\ \hline
			17 & box/bag-related/container & colourful & tools/elongated \\ \hline
			18 & fruit-related & container/household & fluid-related/drink-related \\ \hline
			19 & transportation-related & beautiful/gorgeous & water-related \\ \hline
			20 & clothing & sports equipment & oriented/many things \\ \hline
			21 & protective/medical-related & flying-/sky-related & decay-related/grainy \\ \hline
			22 & frozen treats/drink & fluid-related/drink-related & white \\ \hline
			23 & presentation/display-related & paper-related/flat & coarse pattern/many things \\ \hline
			24 & body part-related & indoor & red \\ \hline
			25 & headwear-related & outdoor & long/thin \\ \hline
			26 & livestock-related & flat/thin & weapon-/danger-related \\ \hline
			27 & communal/public-related & side-by-side/neat/regular & black \\ \hline
			28 & fashion accessory-related & vegetable-related & household \\ \hline
			29 & insect-related/disgusting & wood-related & feminine (stereotypical) \\ \hline
			30 & home appliances & body-parts & body part-related \\ \hline
			31 & food condiment-related & juice/fruit drink & tubular \\ \hline
			32 & childrens toy/plaything-related & weapon-related & music-/hearing-/hobby-related \\ \hline
			33 & plant-related & fine-grained pattern & grid-/grating-related \\ \hline
			34 & metallic/handle-related & insect-related/disgusting & repetitive/spiky \\ \hline
			35 & sea creature-/water-related & grainy & construction-/craftsman-related \\ \hline
			36 & drinkware-related & door-parts/metal & spherical/voluminous \\ \hline
			37 & nursing-/look after-related & underwear/wearable & string-related/stringy \\ \hline
			38 & paper-/fabric-related & entertaining & seating-/standing-/lying-related \\ \hline
			39 & rodlike/elongated & many things/aggregated & flying-/sky-related \\ \hline
			40 & music-related & stacked/clutter-related & disgusting/slimy \\ \hline
			41 & food-/dining-related & fruit-related & elliptical/curved \\ \hline
			42 & cereal/grain-related & baby care-related & sand-colored \\ \hline
			43 & ball sport-related & construction-related & green \\ \hline
			44 & avian-/bird-related & dense/many small things & bathroom-/wetness-related \\ \hline
			45 & mechanical/machinery-related & textile & yellow \\ \hline
			46 & ship/water-related & cute/soft & heat-/light-related \\ \hline
			47 & automobile part-related & precious/jewelry-related & beams-/mesh-related \\ \hline
			48 & hairdressing-/beauty-related & wearing jewelry-related & foot-/walking-related \\ \hline
			49 & sound-making tool-related & leg-/walking-related & box-related/container \\ \hline
			50 & footwear-/foot-related & water-related & stick-shaped/cylindrical \\ \hline
			51 & tree-/wood-related & sea creature-/water-related & head-related \\ \hline
			52 & movement-related/vehicle & wearable & upright/elongated/volumous \\ \hline
			53 & fire/burning-related & summer-related/lite & pointed/spiky \\ \hline
			54 & coldness-related/winter-related & elongated/stick-related & child-related/cute \\ \hline
			55 & nut-related & fire/burning-related & farm-related/historical \\ \hline
			56 & sharp & medical care-related & seeing-related/small/round \\ \hline
			57 & hot drink-related & musical instruments & medicine-related \\ \hline
			58 & game-/board game-related & transportation-related & dessert-related \\ \hline
			59 & high-tech/national defense-related & daily living equipment & orange \\ \hline
			60 & optic-/lens-related & ship-/water-related & thin/flat \\ \hline
			61 & music player-related & ball-/playing-related & cylindrical/conical/cushioning \\ \hline
			62 & disgusting/dirty & game-/amusement-related & coldness-related/winter-related \\ \hline
			63 & thin/flat-related & childrens toy-related & measurement-related/numbers-related \\ \hline
			64 & spiny/spiky/spiculate & round/curvature-related & fluffy/soft \\ \hline
			65 & tableware-related & coarse pattern/many things & masculine (stereotypical) \\ \hline
			66 & string-related/stringy  & birds-related & fine-grained pattern \\ \hline
		\end{tabular}
	}
	\label{table:66dim_labels}
\end{table}

\begin{table}[!htbp]
	\centering
	\caption{Dimension labels summed up by the human experts and the MLLM (here, gemini-pro-1.5-exp, related to Fig. \ref{fig:dim_vis}). MLLM matches human annotation highly consistently marked with {\color{green}$\checkmark\checkmark$}, consistent with {\color{green}$\checkmark$}, and inconsistent with {\color{red}\scriptsize{\XSolidBrush}}. While MLLM excels at concrete comparative tasks (like triplet odd-one-out selection), it shows limitations in dimension naming tasks that require abstracting and generalizing across diverse visual and semantic features.}
	\ra{0.95}
	\co{1.5pt}
	\resizebox{17.5cm}{!}{
		\begin{tabular}{|c|p{4.8cm}|p{6.3cm}|p{4.5cm}|p{6cm}|}
			\hline
			& \multicolumn{2}{|c|}{\textbf{Dimension labels of LLM (GPT3.5-Turbo)}} & \multicolumn{2}{|c|}{\textbf{Dimension labels of  MLLM (Gemini Pro Vision 1.0)}} \\ \hline
			\textbf{Dim.} & \textbf{\ \ Annotated by human experts} & \textbf{\qquad\quad Annotated by MLLM} & \textbf{\ Annotated by human experts} & \textbf{\qquad\quad Annotated by MLLM} \\ \hline
			1 & household-/furnishing-related & household items/home furnishings {\color{green}\checkmark}{\color{green}\checkmark} & multiple/repetitive & secondhand goods/flea market {\color{red}\scriptsize{\XSolidBrush}}\\ \hline
			2 & wild animal-related & wild animals/animals in the wild {\color{green}\checkmark}{\color{green}\checkmark} & animal-related & wild animals {\color{green}\checkmark} \\ \hline
			3 & modern life-related & modes of transportation {\color{red}\scriptsize{\XSolidBrush}} & food-related & breakfast foods/brunch dishes {\color{green}\checkmark} \\ \hline
			4 & support frame-related & simple machines/mechanical advantage {\color{red}\scriptsize{\XSolidBrush}} & plant-related/green & green plants {\color{green}\checkmark}{\color{green}\checkmark} \\ \hline
			5 & electronics/technology & old technology/obsolete technology {\color{green}\checkmark} & grand/spectacular & different watercrafts/vessels {\color{red}\scriptsize{\XSolidBrush}}\\ \hline
			6 & cuisines-/food-related & dishes/food {\color{green}\checkmark}{\color{green}\checkmark} & electronics/technology & electronic devices/obsolete technology {\color{green}\checkmark}{\color{green}\checkmark} \\ \hline
			7 & weapon-related & weapons/weaponry {\color{green}\checkmark}{\color{green}\checkmark} & vehicle-related & modes of transportation/vehicles {\color{green}\checkmark}{\color{green}\checkmark} \\ \hline
			8 & dessert-/sweet-related & sweets/candy {\color{green}\checkmark}{\color{green}\checkmark} & serried/stacked & round objects/circular shapes {\color{red}\scriptsize{\XSolidBrush}}\\ \hline
			9 & amusement-/playing-related & recreational activities/outdoor fun {\color{green}\checkmark} & construction-/craftsman-related & hand tools/tools {\color{green}\checkmark} \\ \hline
			10 & precious/jewelry-related & jewelry \& gems {\color{green}\checkmark}{\color{green}\checkmark} & handtool-related/metallic & household tools {\color{green}\checkmark} \\ \hline
			11 & grainy & raw materials {\color{red}\scriptsize{\XSolidBrush}} & clothing-related/textile & clothing, apparel {\color{green}\checkmark}{\color{green}\checkmark} \\ \hline
			12 & hand tool-related & tools/hand tools {\color{green}\checkmark}{\color{green}\checkmark} & valuable/antique & antique/vintage {\color{green}\checkmark} \\ \hline
			13 & vegetable-related & vegetables/produce {\color{green}\checkmark}{\color{green}\checkmark} & gorgeous/feminine & gifts/presents {\color{red}\scriptsize{\XSolidBrush}}\\ \hline
			14 & danger/harm & death/suffering {\color{red}\scriptsize{\XSolidBrush}} & household-related & furniture/home furnishings {\color{green}\checkmark} \\ \hline
			15 & fluffy-related/lightsome & fall/autumn {\color{red}\scriptsize{\XSolidBrush}}& house-/furniture-related & home furniture {\color{green}\checkmark}{\color{green}\checkmark} \\ \hline
			16 & fastening-/stabilizing-related & craft supplies/crafting materials {\color{red}\scriptsize{\XSolidBrush}} & houseware-related & household appliances/items {\color{green}\checkmark}{\color{green}\checkmark} \\ \hline
			17 & box/bag-related/container & storage/containers {\color{green}\checkmark}{\color{green}\checkmark} & colourful & bright colors {\color{green}\checkmark} \\ \hline
			18 & fruit-related & fruits/fruit varieties {\color{green}\checkmark}{\color{green}\checkmark} & container/household & glass containers/containers for liquids {\color{green}\checkmark} \\ \hline
			19 & transportation-related & modes of transportation/vehicles {\color{green}\checkmark}{\color{green}\checkmark} & beautiful/gorgeous & luxury/wealth {\color{red}\scriptsize{\XSolidBrush}} \\ \hline
			20 & clothing & women's clothing {\color{green}\checkmark} & sports equipment & children's toys/recreational equipment {\color{red}\scriptsize{\XSolidBrush}} \\ \hline
			21 & protective/medical-related & safety equipment/protective gear {\color{green}\checkmark}{\color{green}\checkmark} & flying-/sky-related & air \& space/flight \& aerospace {\color{green}\checkmark} \\ \hline
			22 & frozen treats/drink & desserts/sweet treats {\color{green}\checkmark} & fluid-related/drink-related & beverages/drinks {\color{green}\checkmark}{\color{green}\checkmark} \\ \hline
			23 & presentation/display-related & home improvement {\color{red}\scriptsize{\XSolidBrush}}& paper-related/flat & office supplies / stationery {\color{green}\checkmark} \\ \hline
			24 & body part-related & body parts {\color{green}\checkmark}{\color{green}\checkmark} & indoor & household items/home goods {\color{red}\scriptsize{\XSolidBrush}} \\ \hline
			25 & headwear-related & head coverings/headwear {\color{green}\checkmark}{\color{green}\checkmark} & outdoor & outdoor scenes {\color{green}\checkmark}{\color{green}\checkmark} \\ \hline
			26 & livestock-related & farm animals/livestock {\color{green}\checkmark}{\color{green}\checkmark} & flat/thin & common materials/everyday items {\color{red}\scriptsize{\XSolidBrush}} \\ \hline
			27 & communal/public-related & street furniture/public amenities {\color{green}\checkmark} & side-by-side/neat/regular & storage/supplies/inventory/stock {\color{red}\scriptsize{\XSolidBrush}} \\ \hline
			28 & fashion accessory-related & fashion accessories {\color{green}\checkmark}{\color{green}\checkmark} & vegetable-related & green vegetables/leafy greens {\color{green}\checkmark}{\color{green}\checkmark} \\ \hline
			29 & insect-related/disgusting & insects/bugs {\color{green}\checkmark}{\color{green}\checkmark} & wood-related & wood products/wooden objects {\color{green}\checkmark}{\color{green}\checkmark} \\ \hline
			30 & home appliances & home appliances {\color{green}\checkmark}{\color{green}\checkmark} & body-parts & human body parts {\color{green}\checkmark}{\color{green}\checkmark} \\ \hline
			31 & food condiment-related & food spreads/condiments {\color{green}\checkmark}{\color{green}\checkmark} & juice/fruit drink & food and drinks {\color{green}\checkmark} \\ \hline
			32 & childrens toy/plaything-related & children's toys/playthings {\color{green}\checkmark}{\color{green}\checkmark} & weapon-related & weapons/explosives {\color{green}\checkmark}{\color{green}\checkmark} \\ \hline
			33 & plant-related & flowers/plants {\color{green}\checkmark}{\color{green}\checkmark} & fine-grained pattern & raw materials/natural resources {\color{red}\scriptsize{\XSolidBrush}} \\ \hline
			34 & metallic/handle-related & door hardware/door parts {\color{green}\checkmark} & insect-related/disgusting & insects and invertebrates {\color{green}\checkmark}{\color{green}\checkmark} \\ \hline
			35 & sea creature-/water-related & seafood/sea creatures {\color{green}\checkmark} & grainy & natural materials/raw materials {\color{red}\scriptsize{\XSolidBrush}} \\ \hline
			36 & drinkware-related & household items/containers. {\color{green}\checkmark} & door-parts/metal & door hardware/door parts {\color{green}\checkmark}{\color{green}\checkmark} \\ \hline
			37 & nursing-/look after-related & furniture/household items {\color{red}\scriptsize{\XSolidBrush}}& underwear/wearable & women's apparel/lingerie {\color{green}\checkmark} \\ \hline
			38 & paper-/fabric-related & household items {\color{red}\scriptsize{\XSolidBrush}} & entertaining & analog technology/vintage items {\color{red}\scriptsize{\XSolidBrush}}\\ \hline
			39 & rodlike/elongated & long/thin/rod-like objects {\color{green}\checkmark}{\color{green}\checkmark} & many things/aggregated & industrial supplies {\color{red}\scriptsize{\XSolidBrush}} \\ \hline
			40 & music-related & musical instruments {\color{green}\checkmark}{\color{green}\checkmark} & stacked/clutter-related & garage sale/flea market {\color{red}\scriptsize{\XSolidBrush}} \\ \hline
			41 & food-/dining-related & party food/snacks {\color{green}\checkmark} & fruit-related & fruits/fruit variety {\color{green}\checkmark}{\color{green}\checkmark} \\ \hline
			42 & cereal/grain-related & wheat products/foods made from wheat {\color{green}\checkmark}{\color{green}\checkmark} & baby care-related & baby items/baby products {\color{green}\checkmark}{\color{green}\checkmark} \\ \hline
			43 & ball sport-related & sports equipment {\color{green}\checkmark}{\color{green}\checkmark} & construction-related & urban infrastructure/city elements {\color{red}\scriptsize{\XSolidBrush}} \\ \hline
			44 & avian-/bird-related & birds/bird species {\color{green}\checkmark}{\color{green}\checkmark} & dense/many small things & natural textures/organic materials {\color{red}\scriptsize{\XSolidBrush}} \\ \hline
			45 & mechanical/machinery-related & simple machines/mechanical parts {\color{green}\checkmark}{\color{green}\checkmark} & textile & household linens/fabric goods {\color{green}\checkmark} \\ \hline
			46 & ship/water-related & boats/watercraft {\color{green}\checkmark}{\color{green}\checkmark} & cute/soft & handmade/crafts {\color{red}\scriptsize{\XSolidBrush}}\\ \hline
			47 & automobile part-related & car parts {\color{green}\checkmark}{\color{green}\checkmark} & precious/jewelry-related & valuable possessions/precious items {\color{green}\checkmark} \\ \hline
			48 & hairdressing-/beauty-related & hair styling/hair care {\color{green}\checkmark}{\color{green}\checkmark} & wearing jewelry-related & hair accessories {\color{red}\scriptsize{\XSolidBrush}} \\ \hline
			49 & sound-making tool-related & musical instruments/sound makers {\color{green}\checkmark}{\color{green}\checkmark} & leg-/walking-related & human legs {\color{green}\checkmark}{\color{green}\checkmark} \\ \hline
			50 & footwear-/foot-related & footwear/shoes {\color{green}\checkmark}{\color{green}\checkmark} & water-related & water activities/aquatic life {\color{green}\checkmark} \\ \hline
			51 & tree-/wood-related & trees/wood {\color{green}\checkmark}{\color{green}\checkmark} & sea creature-/water-related & fishing/seafood {\color{red}\scriptsize{\XSolidBrush}} \\ \hline
			52 & movement-related/vehicle & recreational vehicles/personal vehicles {\color{green}\checkmark} & wearable & personal accessories/personal items {\color{green}\checkmark} \\ \hline
			53 & fire/burning-related & fire/sources of fire {\color{green}\checkmark}{\color{green}\checkmark} & summer-related/lite & beach vacation/swimming gear {\color{red}\scriptsize{\XSolidBrush}} \\ \hline
			54 & coldness-related/winter-related & winter/cold weather {\color{green}\checkmark}{\color{green}\checkmark} & elongated/stick-related & handmade crafts/diy projects {\color{red}\scriptsize{\XSolidBrush}}\\ \hline
			55 & nut-related & nuts/edible nuts {\color{green}\checkmark}{\color{green}\checkmark} & fire/burning-related & fire/combustion {\color{green}\checkmark}{\color{green}\checkmark} \\ \hline
			56 & sharp & sharp objects/blades {\color{green}\checkmark}{\color{green}\checkmark} & medical care-related & assistive devices/mobility aids {\color{green}\checkmark} \\ \hline
			57 & hot drink-related & coffee \& tea {\color{green}\checkmark}{\color{green}\checkmark} & musical instruments & musical instruments {\color{green}\checkmark}{\color{green}\checkmark} \\ \hline
			58 & game-/board game-related & board games/indoor games {\color{green}\checkmark}{\color{green}\checkmark} & transportation-related & vehicles/motor vehicles {\color{green}\checkmark}{\color{green}\checkmark} \\ \hline
			59 & high-tech/national defense-related & vehicles/transportation {\color{red}\scriptsize{\XSolidBrush}} & daily living equipment & hotel linens/hotel supplies {\color{red}\scriptsize{\XSolidBrush}} \\ \hline
			60 & optic-/lens-related & optical lenses/optics {\color{green}\checkmark}{\color{green}\checkmark} & ship-/water-related & sea vessels/watercraft {\color{green}\checkmark}{\color{green}\checkmark} \\ \hline
			61 & music player-related & music players/audio devices {\color{green}\checkmark}{\color{green}\checkmark} & ball-/playing-related & sports equipment {\color{green}\checkmark} \\ \hline
			62 & disgusting/dirty & water infrastructure/urban utilities {\color{red}\scriptsize{\XSolidBrush}}& game-/amusement-related & sports/games {\color{green}\checkmark} \\ \hline
			63 & thin/flat-related & shiny materials/reflective surfaces {\color{red}\scriptsize{\XSolidBrush}} & childrens toy-related & children's toys {\color{green}\checkmark}{\color{green}\checkmark} \\ \hline
			64 & spiny/spiky/spiculate & sharp objects/pointy things {\color{green}\checkmark}{\color{green}\checkmark} & round/curvature-related & obsolete technology {\color{red}\scriptsize{\XSolidBrush}}\\ \hline
			65 & tableware-related & kitchen utensils {\color{green}\checkmark}{\color{green}\checkmark} & coarse pattern/many things & arts and crafts {\color{red}\scriptsize{\XSolidBrush}}\\ \hline
			66 & string-related/stringy & knots and cords {\color{green}\checkmark} & birds-related & birds/bird species {\color{green}\checkmark}{\color{green}\checkmark} \\ \hline
		\end{tabular}
	}
	\label{table:66dim_labels_artificial_vs_automatic}
\end{table}

\begin{figure}[!htbp]
	\centering
    \includegraphics[width=17cm]{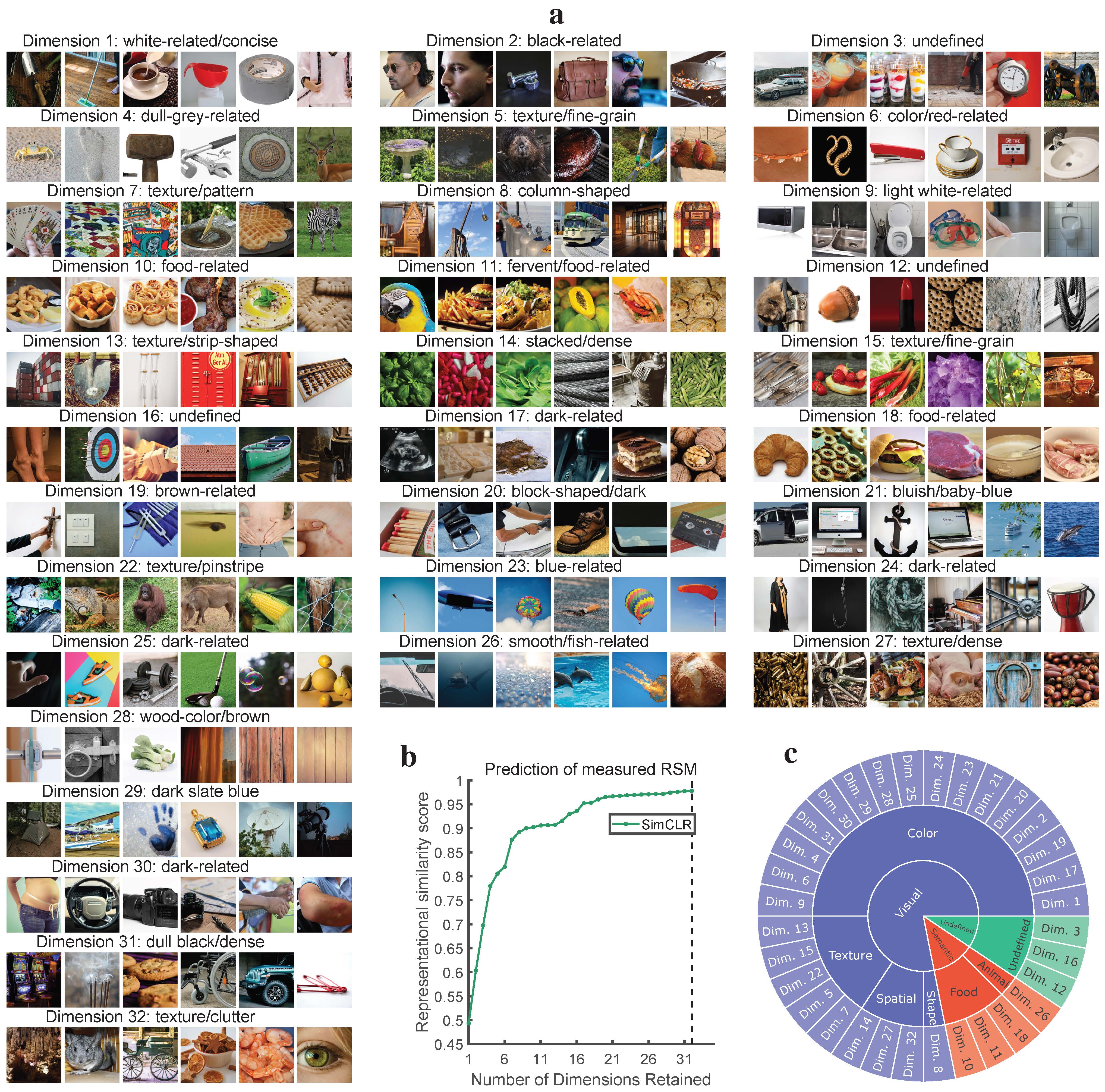}
	\caption{\textbf{Object dimensions (1-32) illustrating their interpretability for self-supervised learning model SimCLR (related to Fig. \ref{fig:dim_vis}).} \textbf{a}, Each dimension is illustrated with the top 6 images with the highest weights along this dimension. \textbf{b}, Dimensions retained by SimCLR and the ability to predict its behavioral RSMs. \textbf{c}, Attribution of the 32 dimensions of the SimCLR model, where the visual dimensions occupy the vast majority, and only a few semantic dimensions.}
	\label{fig:dim_vis_simclr}
\end{figure}

\begin{figure}[!htbp]
	\centering
    \includegraphics[width=17cm]{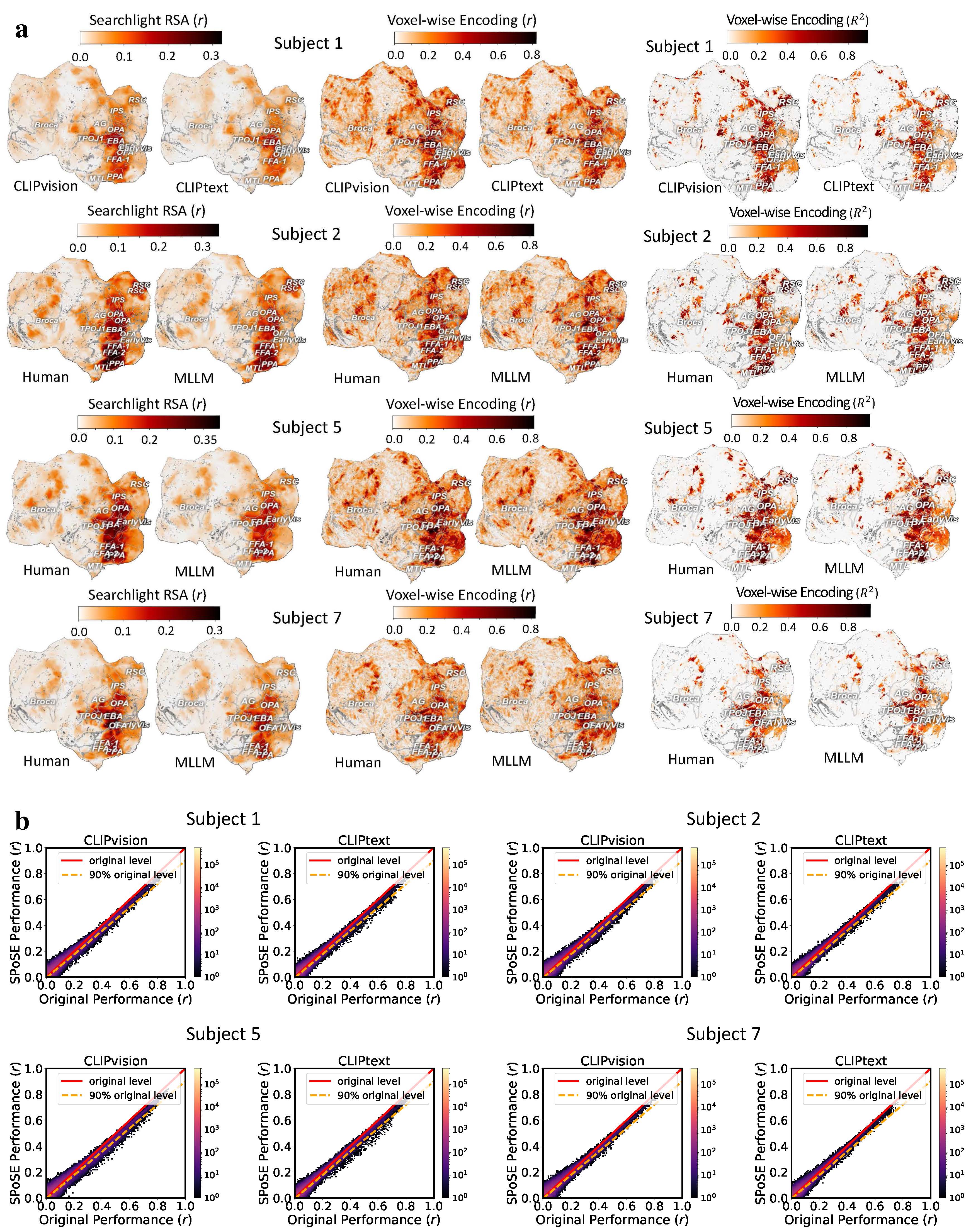}
	\caption{\textbf{More results on the relationship between model and brain representations (related to Fig. \ref{fig:rsa}).} \textbf{a}, Flattened cortical maps for more models and subjects. Performance was evaluated by using both Pearson's correlation ($r$) and the noise-normalized $R^2$. \textbf{b}, Voxel-wise encoding performance using the original high-dimensional model features and the low-dimensional SPoSE embeddings of CLIP model.}
	\label{fig:rsa_s2_s8}
\end{figure}

	\newpage
	\section*{Supplementary information}
	\beginsupplement
	\supplementfigure	
	\supplementtable	
	
	\begin{figure}[!htbp]
	\centering
	\includegraphics[width=17cm]{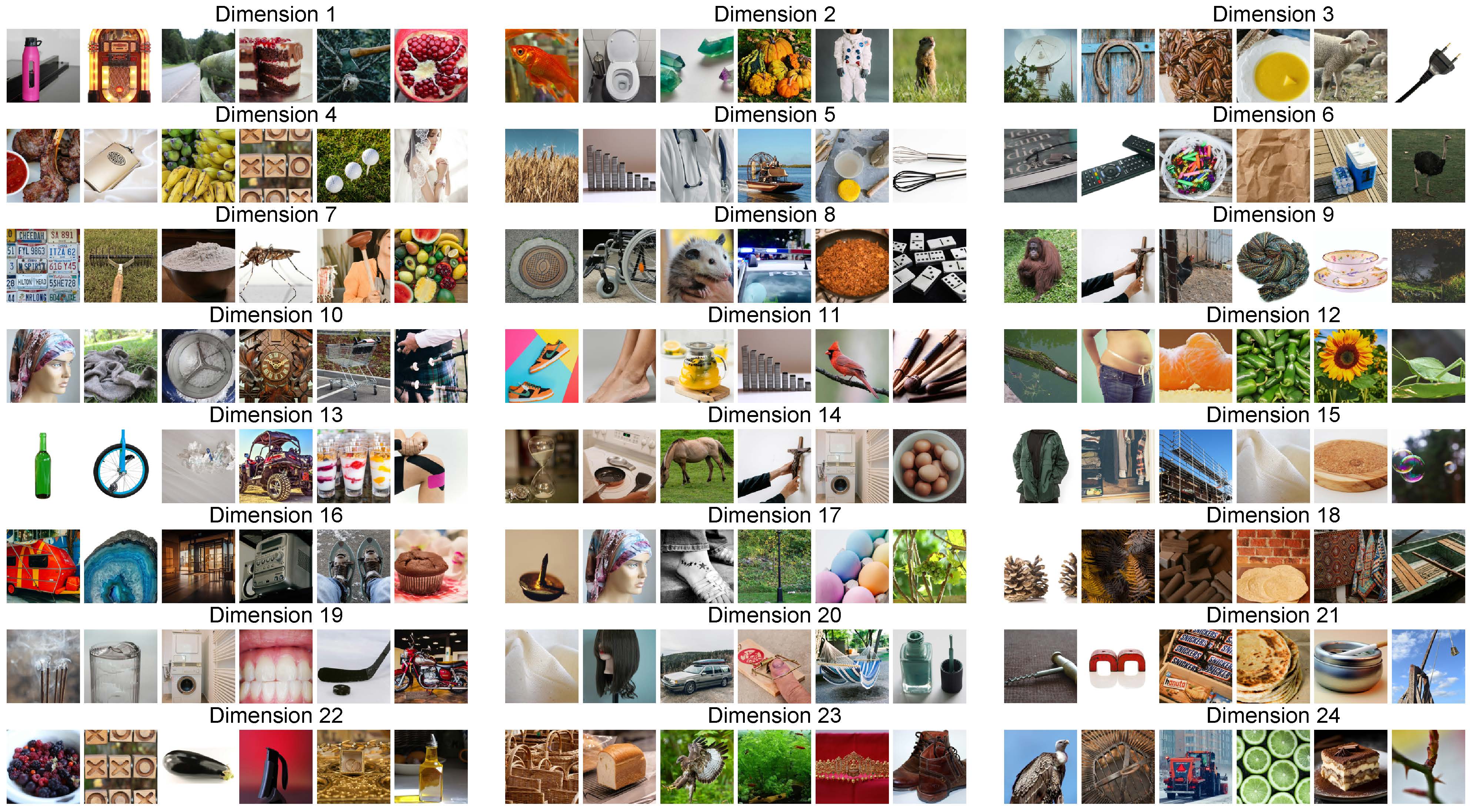}
	\caption{ \textbf{Top 24 dimensions for "random representation" model (related to Fig. 4).} We constructed representations of the 1,854 object concepts using 1,000-dimensional random vectors, generated 4.7 million odd-one-out data points based on cosine distances, and then applied the SPoSE method to learn low-dimensional embeddings. Each dimension was illustrated with the top 6 images with the highest weights along this dimension. These dimensions exhibit no interpretability whatsoever. This strongly suggests that the interpretability of the dimensions obtained from LLM/MLLM is primarily attributable to the models' representations rather than the SPoSE method itself. For this figure, all images were replaced by images with similar appearance from the public domain. Images used under a CC0 license, from Pixabay and Pexels.}
	\label{fig:dim_vis_random}
	
\end{figure}

\begin{figure}[!htbp]
	\centering
	\includegraphics[scale=0.5]{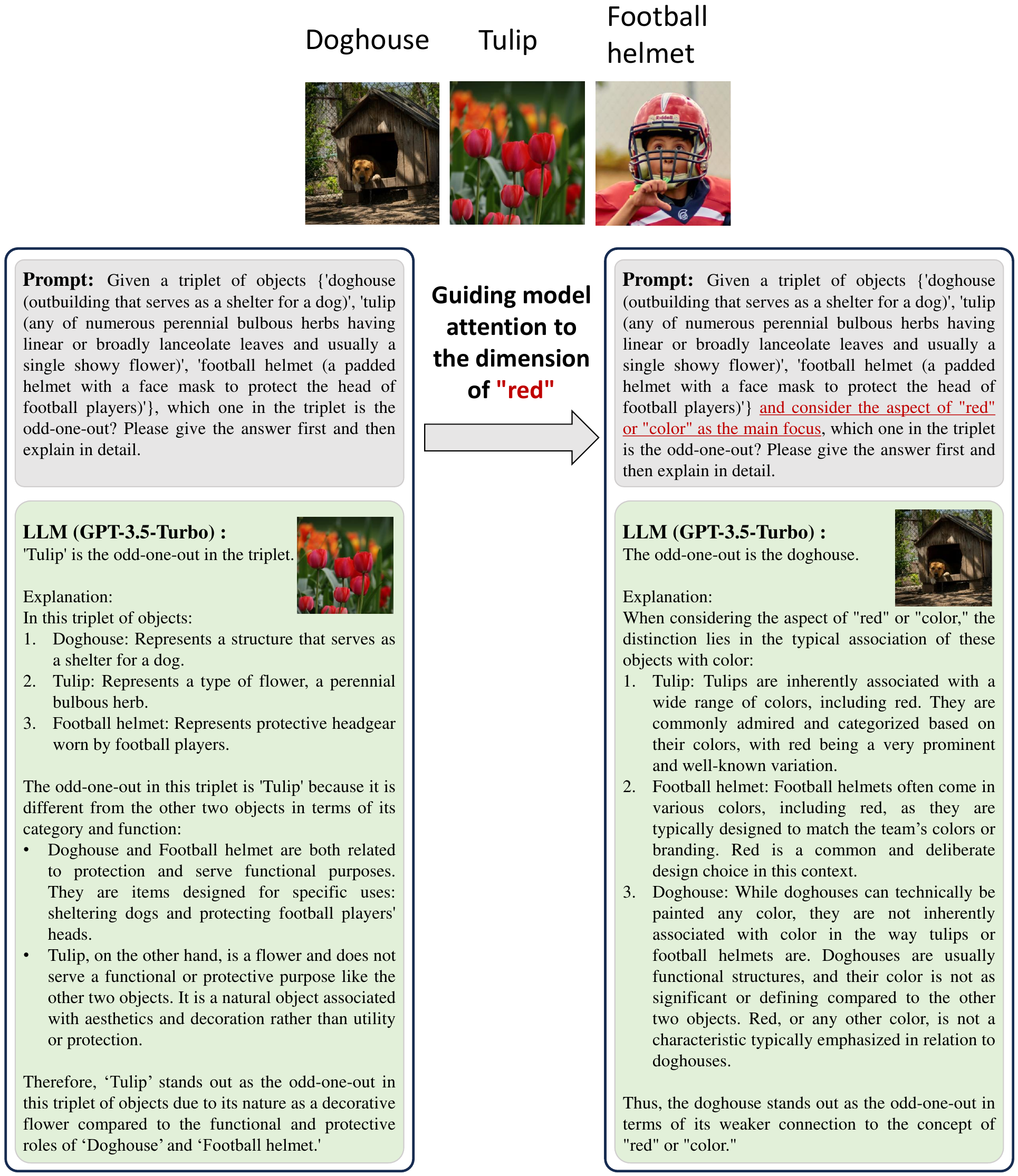}
	\caption{\textbf{Guiding LLM's attention to the target dimension by using tailored prompts (related to Fig. 5).} We added the phrase "consider the aspect of "red" or "color" as the main focus" to the prompt of LLM. As can be seen, when the prompt included guidance on the dimensions prioritized by humans ("red"), the LLM was able to make choice consistent with human judgment. For this figure, all images were replaced by images with similar appearance from the public domain. Images used under a CC0 license, from Pixabay and Pexels.}
	\label{fig:guiding_llm}
\end{figure}

\begin{figure}[!htbp]
	\centering
	\includegraphics[scale=0.5]{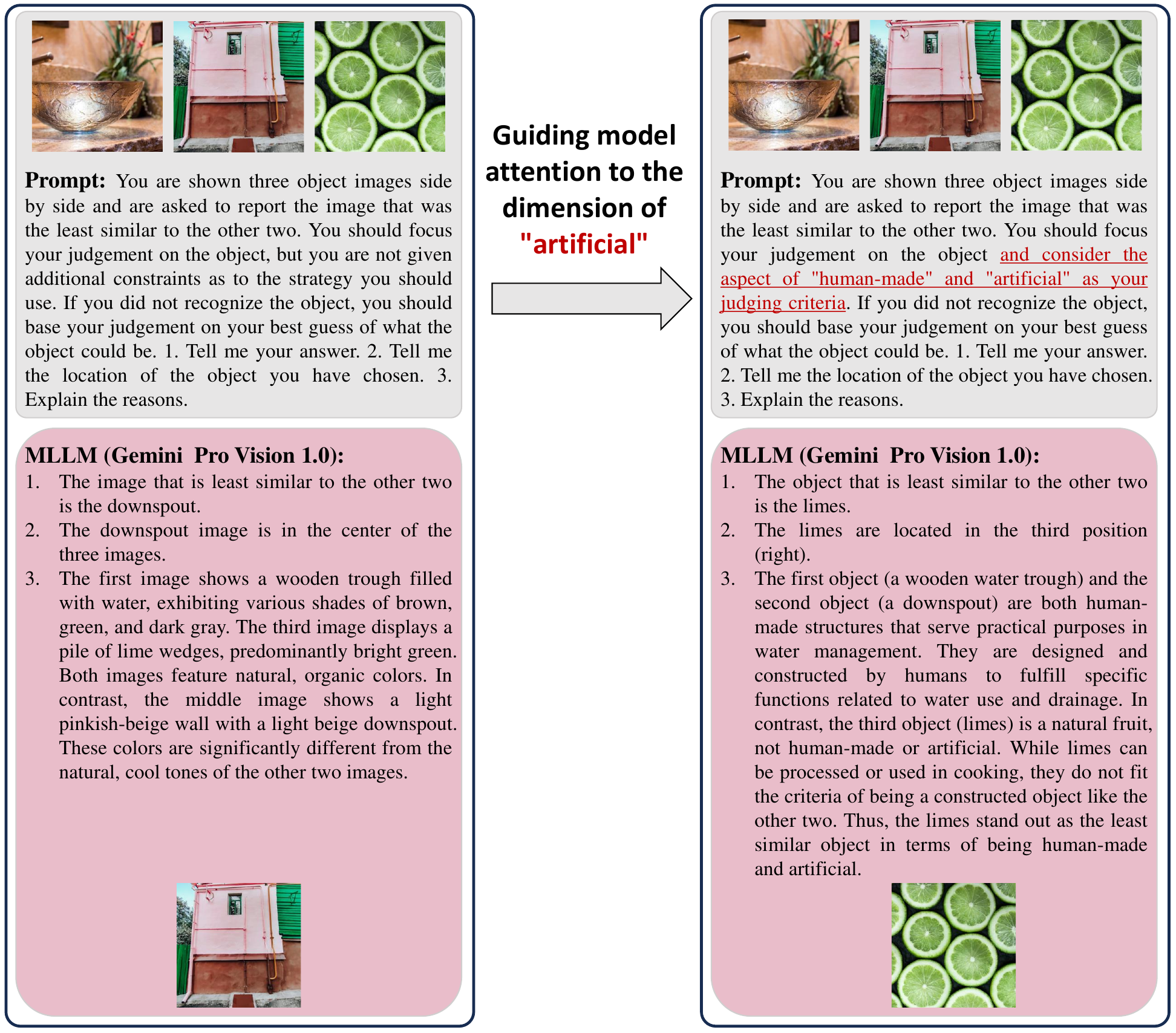}
	\caption{\textbf{Guiding MLLM's attention to the target dimension by using tailored prompts (related to Fig. 5).}  We added the phrase "consider the aspect of "human-made" and "artificial" as your judging criteria" to the prompt of MLLM. As can be seen, when the prompt included guidance on the dimensions prioritized by humans ("artificial"), the MLLM was able to make choice consistent with human judgment. For this figure, all images were replaced by images with similar appearance from the public domain. Images used under a CC0 license, from Pixabay and Pexels.}
	\label{fig:guiding_mllm}
\end{figure}

\begin{figure}[!htbp]
	\centering
	\begin{tikzpicture}
		\node[anchor=south east] at (-7,11.6) {\includegraphics[scale=0.8]{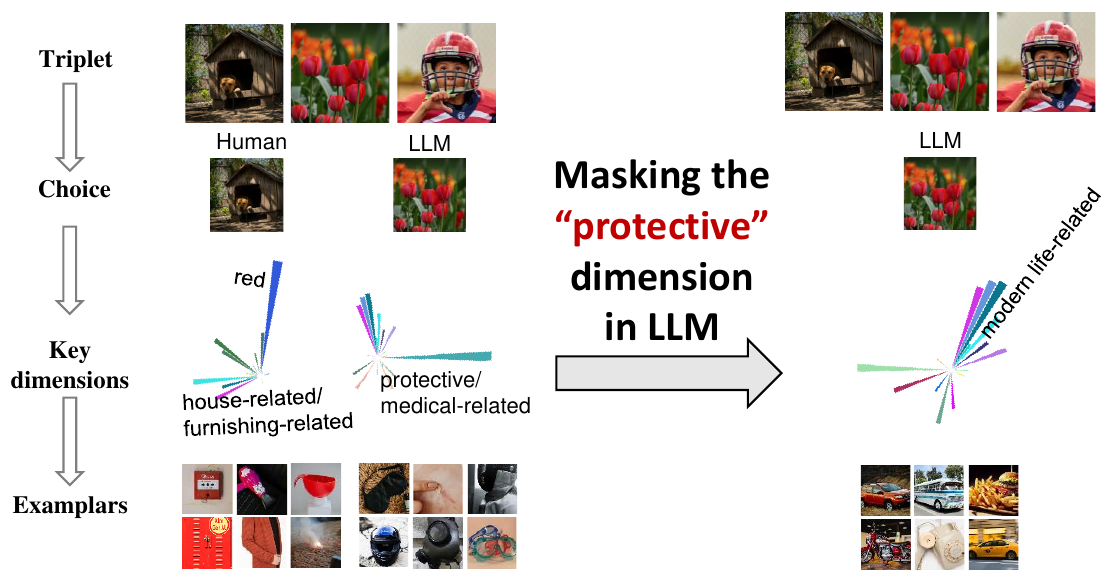}};
		\node at (-22.5,19.2){\textbf{\Large a}};
		\node at (-22.5,10.2){\textbf{\Large b}};	
		\node[anchor=south east] at (-7.3,3) {\includegraphics[scale=0.75]{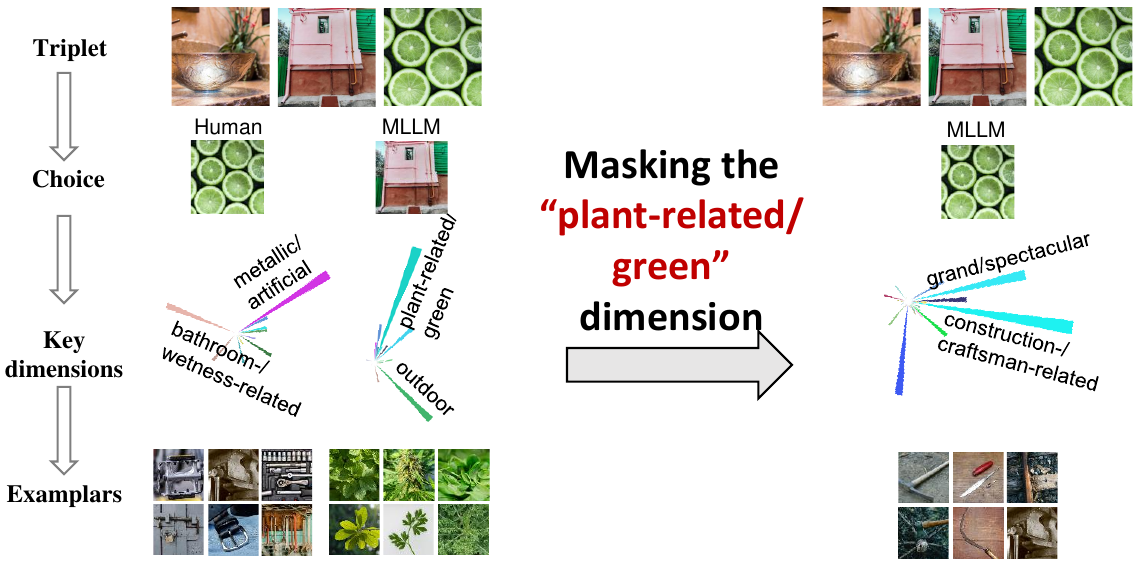}};	
	\end{tikzpicture}
	\caption{\textbf{Masking the most critical dimension currently prioritized by the model but deviating from human preferences (related to Fig. 5).} \textbf{a}, After masking the "protective" dimension, the LLM's odd-one-out choice using the remaining 65 dimensions remained unchanged, but the key dimension it relied on shifted to "modern life-related." \textbf{b}, After masking the "plant-related/green" dimension, the MLLM's choice changed from "downspout" to "limes," and the key dimension it relied on shifted to "construction-/craftsman-related." From these two examples, it can be seen that directly masking certain key dimensions of the LLM/MLLM may or may not change the model's behavioral choices. This intervention method has poor controllability over the model's behavioral choices and the key dimensions it relies on, making it difficult to ensure that the model's choices and the dimensions it relies on will become more aligned with human judgments. For this figure, all images were replaced by images with similar appearance from the public domain. Images used under a CC0 license, from Pixabay and Pexels.}
	\label{fig:masking_llm_mllm}
\end{figure}

\begin{figure}[!htbp]
	\centering
	\includegraphics[scale=0.6]{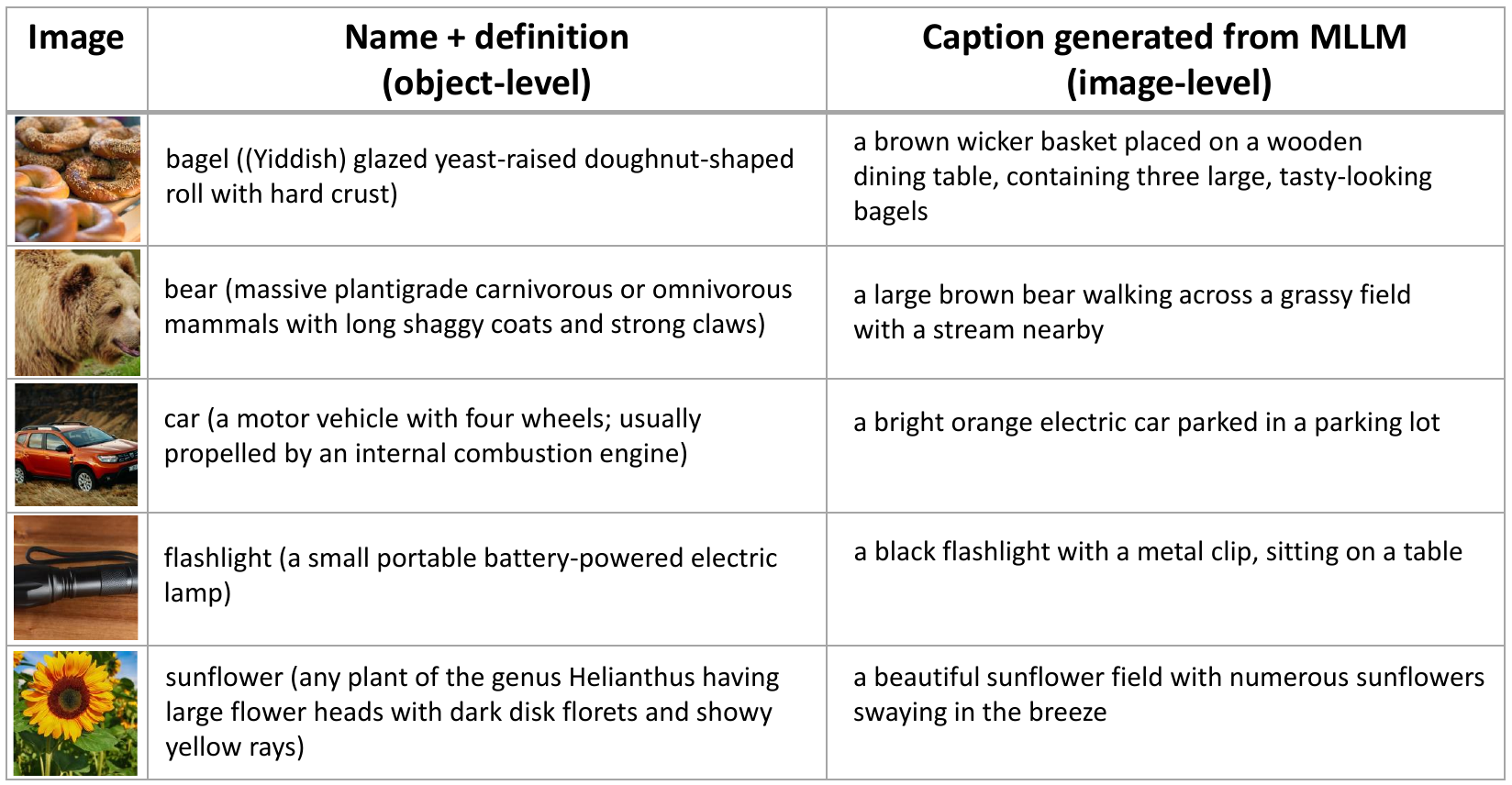}
	\caption{ \textbf{Two kinds of textual descriptions for example images (related to Fig. 1).} \\ \textbf{Object-level annotations}: These annotations focus on the abstract, categorical representation of objects, typically using object names and definitions. They are well-suited for probing high-level conceptual understanding and are less sensitive to visual variations within a category. In our study, the LLM experiments using category-based annotations can be viewed as an "object-level" analysis, as they primarily assess the model's ability to distinguish between objects based on their conceptual categories. \\
		\textbf{Image-level annotations}: Here, the MLLM used for image caption generation was LLaVA-13B-v1-1 with the prompt as "Generate a detailed textual description of the image." These annotations capture detailed visual attributes of individual images, such as color, texture, and spatial relationships. They are more appropriate for tasks that require fine-grained visual discrimination or analysis of within-category variations. In our study, the MLLM experiments, which directly process the visual content of images, can be viewed as an "image-level" analysis, as they assess the model's ability to distinguish objects based on their visual features. For this figure, all images were replaced by images with similar appearance from the public domain. Images used under a CC0 license, from Pixabay and Pexels.}
	\label{fig:image_caption_examples}
\end{figure}

\begin{figure}[!htbp]
	\centering
	\includegraphics[width=17cm]{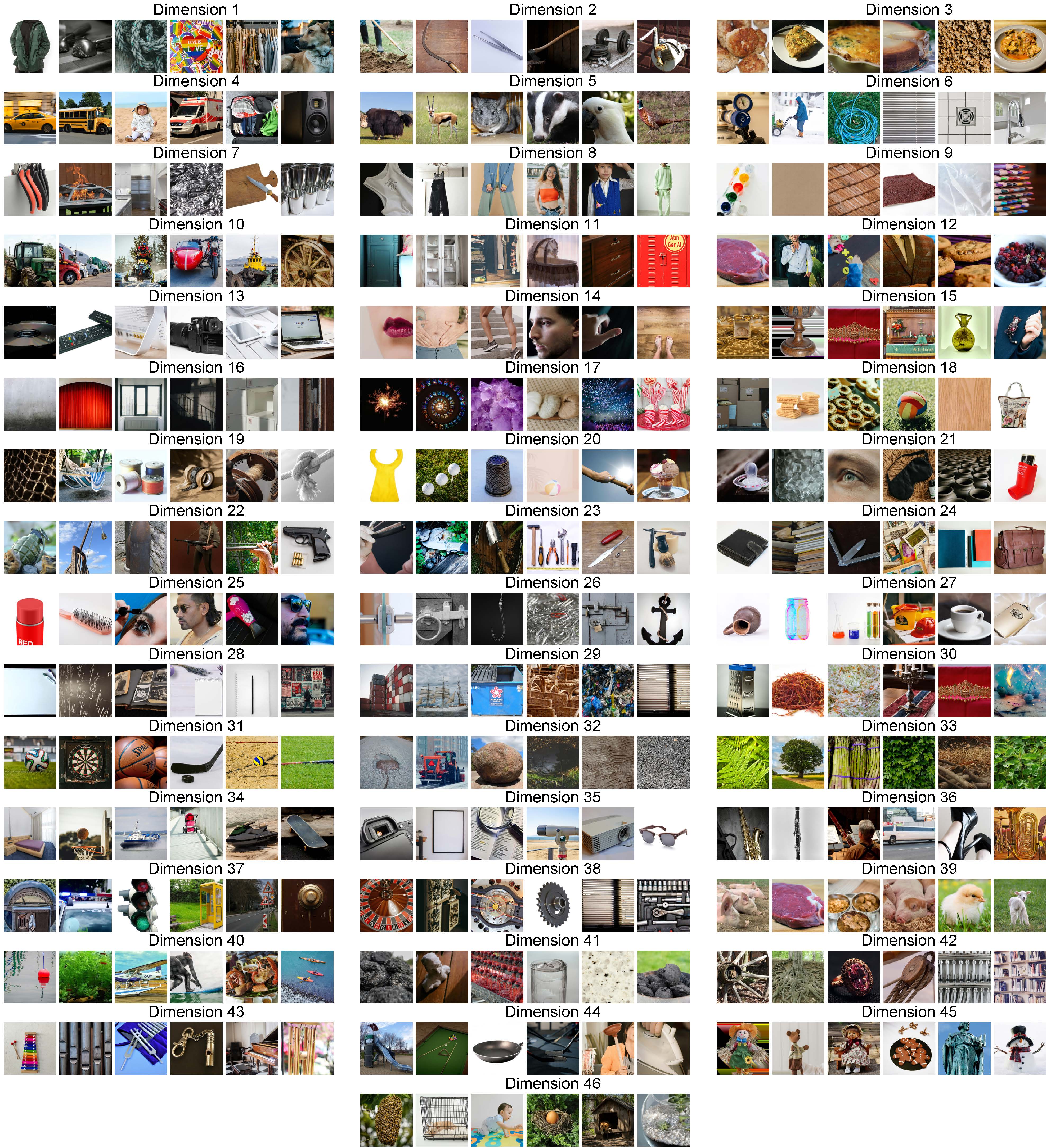}
	\caption{\textbf{Object dimensions (1-46) illustrating their interpretability for LLama3.1 with object-level annotations (related to Fig. 4).} We extracted representations from the object-level descriptions and efficiently constructed 4.7 million odd-one-out triplets based on their cosine distance. We then applied the SPoSE method to learn low-dimensional embeddings from these data, and each dimension was illustrated with the top 6 images with the highest weights along this dimension. For this figure, all images were replaced by images with similar appearance from the public domain. Images used under a CC0 license, from Pixabay and Pexels.}
	\label{fig:dim_vis_LLama3-1-8B}
	
\end{figure}

\begin{figure}[!htbp]
	\centering
	\includegraphics[width=17cm]{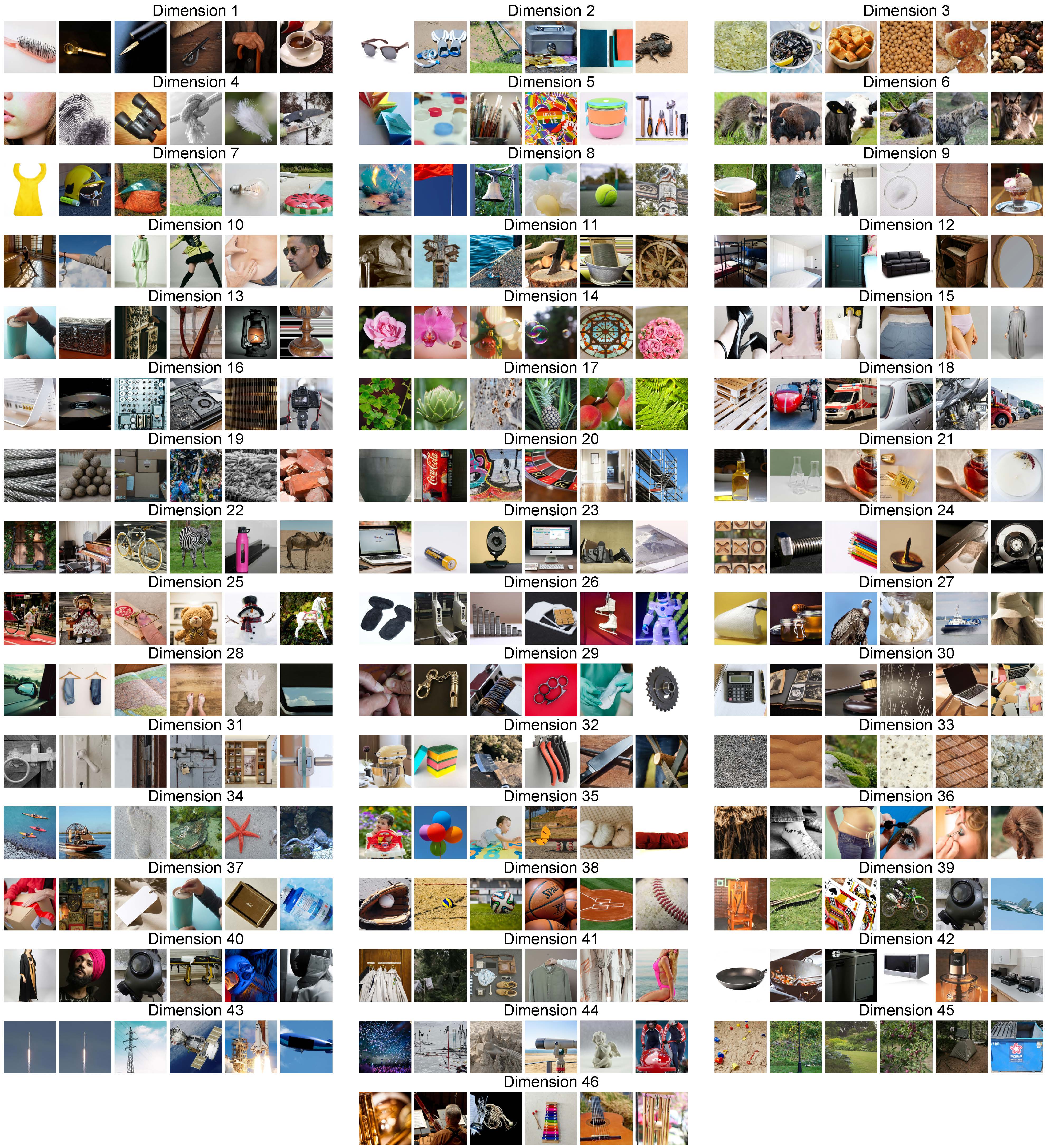}
	\caption{\textbf{Object dimensions (1-46) illustrating their interpretability for LLama3.1 with image-level annotations (related to Fig. 4).} We extracted representations from the image-level descriptions and efficiently constructed 4.7 million odd-one-out triplets based on their cosine distance. We then applied the SPoSE method to learn low-dimensional embeddings from these data, and each dimension was illustrated with the top 6 images with the highest weights along this dimension. In contrast to object-level approach, image-level approach resulted in the emergence of dimensions related to spatial (e.g., Dims. 3, 5, 19), textual (e.g., Dim. 33) and color (e.g., Dim. 14) attributes. For this figure, all images were replaced by images with similar appearance from the public domain. Images used under a CC0 license, from Pixabay and Pexels.}
	\label{fig:dim_vis_LLama3-1-8B_llava_captions}
\end{figure}

\begin{figure}[!htbp]
	\centering
	\includegraphics[scale=0.53]{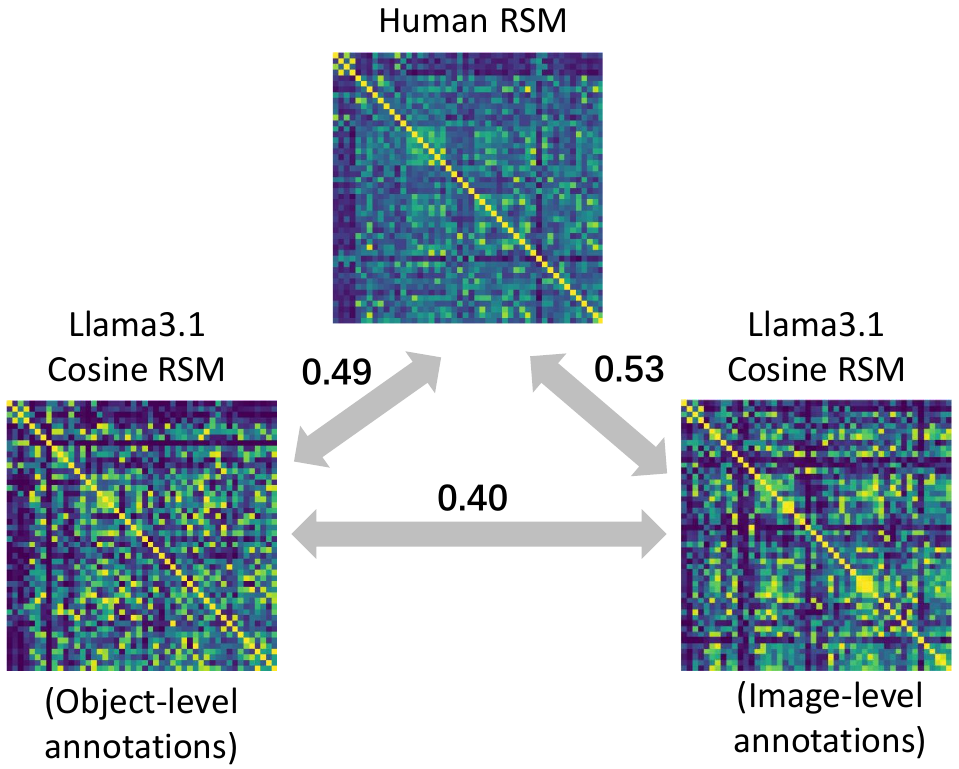}
	\caption{\textbf{Comparison of the RSMs on the 48 typical objects measured by using different image annotation approaches (object-level vs. image-level) (related to Fig. 4).} Cosine RSM was calculated from the model's cosine distance-based odd-one-out data. The numbers on the gray arrows represent the Pearson correlation between different RSM pairs. As can be seen, the RSM corresponding to the image-level annotation method aligns more closely with human judgments (0.53 vs. 0.49), primarily due to the fact that this annotation method leverages a vision-language model to generate image descriptions (effectively providing it with "eyes").}
	\label{fig:object_level_image_level}
\end{figure}

\end{document}